\documentclass{article} 
\usepackage{iclr2026_conference,times}

\usepackage{amsmath,amsfonts,bm}

\def\eqref#1{equation~\ref{#1}}

\def\1{\bm{1}}

\DeclareMathAlphabet{\mathsfit}{\encodingdefault}{\sfdefault}{m}{sl}
\SetMathAlphabet{\mathsfit}{bold}{\encodingdefault}{\sfdefault}{bx}{n}

\usepackage{hyperref}
\usepackage{comment}
\usepackage{pifont} 
\usepackage{array}
\usepackage{graphicx}
\usepackage{wrapfig}
\usepackage{enumitem} 
\usepackage{calc}
\usepackage{booktabs}
\usepackage{xspace}
\usepackage[strict]{changepage} 

\newcommand{\cmark}{\ding{51}} 
\newcommand{\xmark}{\ding{55}} 

\newcommand{\shortname}{TTM\xspace}
\newcommand{\shortnamesvd}{TTM$_{\text{SVD}}$\xspace}
\newcommand{\shortnamecog}{TTM$_{\text{Cog}}$\xspace}

\newcommand{\longname}{Time-to-Move\xspace}

\title{%
\makebox[\textwidth][c]{%
  \parbox{0.65\paperwidth}{\centering
  Time-to-Move: Training-Free Motion Controlled Video Generation via \\ Dual-Clock Denoising
}}}

\author{%
\\
\makebox[\textwidth][c]{%
\begin{tabular}{c}
\textbf{Assaf Singer}$^{1*}$ \quad
\textbf{Noam Rotstein}$^{1*}$ \quad
\textbf{Amir Mann}$^{1}$ \quad
\textbf{Ron Kimmel}$^{1}$ \quad
\textbf{Or Litany}$^{1,2}$ \\
{\normalfont $^{1}$ Technion -- Israel Institute of Technology \quad $^{2}$ NVIDIA}\\
{\normalfont $^{*}$ Equal contribution}
\end{tabular}}%
}

\newcommand{\assaf}[1]{\textcolor{teal}{AS: #1}}

\newcommand{\blue}[1]{\textcolor{blue}{#1}}

\newcommand{\MethodName}{Ctrl+Vid\xspace}

\newif\ifcomments
\commentstrue            

\ifcomments
  \newcommand{\comm}[2][]{\textcolor{red}{\bfseries[#1: #2]}}
\else
  \newcommand{\comm}[2][]{}
\fi

\iclrfinalcopy 
\begin{document}

\maketitle
\vspace{-2pt}

\begin{figure}[h]
  \centering
  \vspace{-12pt}
  \includegraphics[width=1.0\textwidth]{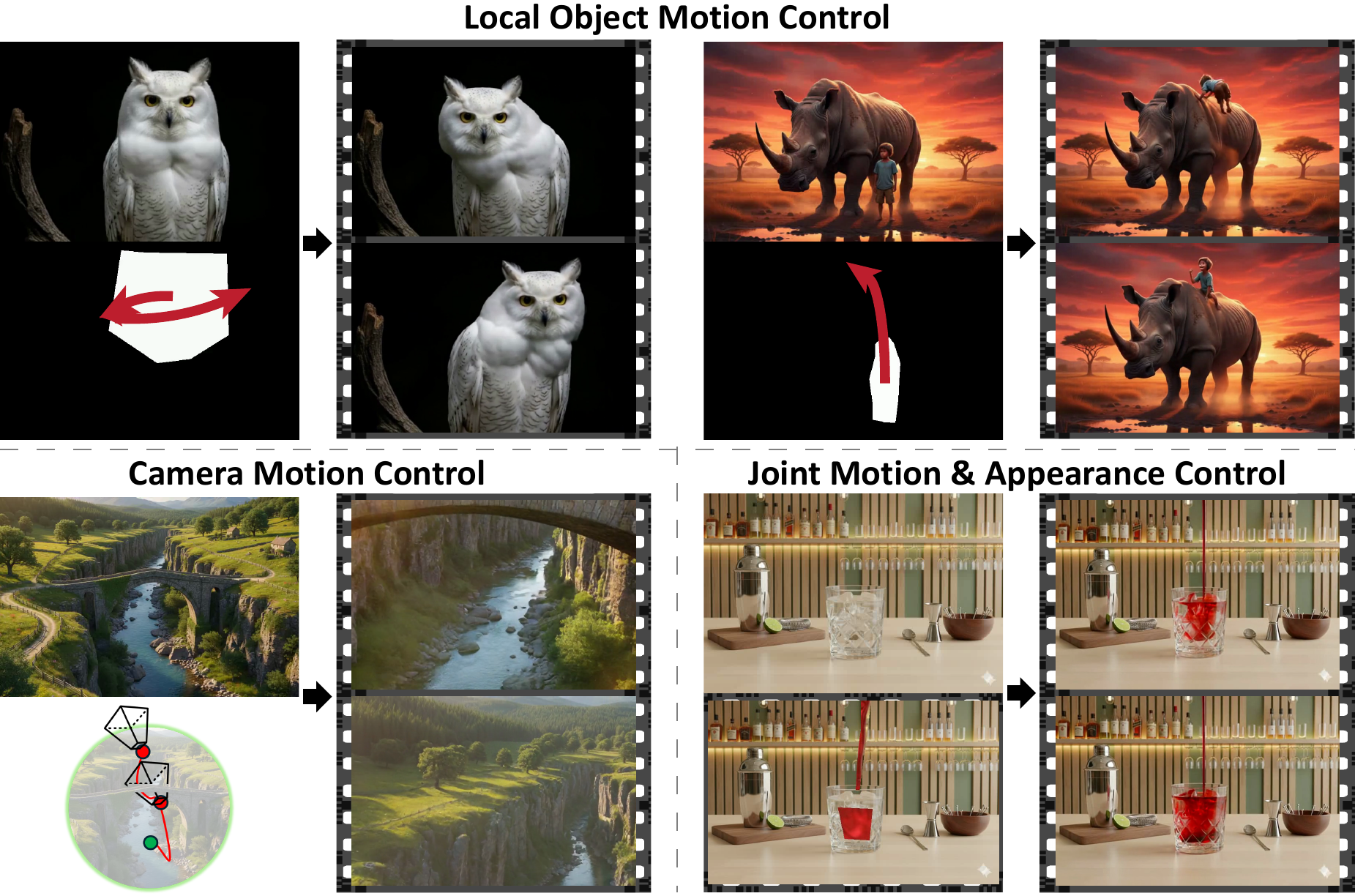}
  \vspace{-18pt}
  \caption{
  Qualitative results of \longname on various tasks.
    \vspace{-8pt}
  }
  \label{fig:teaser}
\end{figure}

\begin{abstract}
\begin{adjustwidth}{-0.2in}{-0.2in}
\vspace{-8pt}
Diffusion-based video generation can create realistic videos, yet existing image- and text-based conditioning fails to offer precise motion control.
Prior methods for motion-conditioned synthesis typically require model-specific fine-tuning, which is computationally expensive and restrictive.
We introduce \longname(\shortname), a training-free, plug-and-play framework for motion- and appearance-controlled video generation with image-to-video (I2V) diffusion models. 
Our key insight is to use crude reference animations obtained through user-friendly manipulations such as cut-and-drag or depth-based reprojection.
Motivated by \emph{SDEdit}’s use of coarse layout cues for image editing, we treat the crude animations as coarse motion cues and adapt the mechanism to the video domain. We preserve appearance with image conditioning and introduce \emph{dual-clock denoising}, a region-dependent strategy that enforces strong alignment in motion-specified regions while allowing flexibility elsewhere, balancing fidelity to user intent with natural dynamics.
This lightweight modification of the sampling process incurs no additional training or runtime cost and is compatible with any backbone. 
Extensive experiments on object and camera motion benchmarks show that \shortname matches or exceeds existing training-based baselines in realism and motion control.
Beyond this, \shortname introduces a unique capability: precise appearance control through pixel-level conditioning, exceeding the limits of text-only prompting.
Visit our \href{https://time-to-move.github.io/}{\blue{project page}} for video examples and code.
\end{adjustwidth}
\end{abstract}

\section{Introduction}
\label{sec:intro}

Diffusion-based video generators have recently achieved remarkable visual quality, yet their controllability remains limited.
Image-to-video (I2V) models partially alleviate this limitation by conditioning on a single input frame, which gives users direct control over the appearance of the generated video.
However, \emph{motion} control remains largely prompt-driven which is often unreliable, coarse, and insufficiently fine-grained for interactive use.
To address this gap, 
 a practical generative video system should provide an interface that defines both \emph{what} moves and \emph{where} it moves, ensuring realistic, temporally coherent motion while preserving the appearance of the input image.
Such fine-grained control would enable interactive content authoring, post-production, and animation prototyping, where creators require precise, local adjustments with fast feedback.
%
%
Existing approaches for controllable motion in generation typically encode user intent through auxiliary control signals such as optical flow or point-trajectories and then \emph{heavily fine-tune} a generator to ingest this motion conditioning \citep{burgert2025go,geng2025motion}. 
Such methods are computationally expensive to train, often compromise the quality of the original model, and remain model-specific, requiring architectural modifications to incorporate the controls. 
This motivates a framework that can be applied to \emph{off-the-shelf} video diffusion backbones without expensive tuning or additional data.



We introduce \longname (\shortname), a \emph{training-free}, architecture-agnostic, plug-and-play inference procedure for video diffusion models that matches the speed of standard generation.
We observe that crude animation inputs, such as those created by simple cut-and-drag manipulation or by straightforward reprojection of the image into novel views using estimated monocular depth, can serve as a useful proxy for the intended target.
Such references capture coarse structure and convey the desired motion, while remaining easy to produce and flexible enough to be made as specific or detailed as the user intends.
To transform these crude signals into realistic videos, we draw on \emph{SDEdit}~\citep{meng2021sdedit}, which shows that coarse structure can be imposed by adding noise to the timestep where the layout is determined.
By analogy, we hypothesize that noising the synthetic reference video to the point where \emph{motion} is established by the video diffusion model can embed the intended dynamics. 
Indeed, this strategy successfully injects motion, but at this noise level fidelity to the reference appearance is lost. 
To mitigate this, we turn to \emph{image}-conditioned video diffusion models, which preserve the identity and scene details of the initial frame and thereby maintain appearance consistency throughout the generated sequence.

Even with appearance preserved, motion cues remain uneven across regions. 
In the synthetic guiding reference video, some regions, such as dragged objects, may contain strong user-specified dynamics, while others remain unspecified. 
These unconstrained regions are not meant to stay static, but rather to adapt naturally in support of the intended movement. 
To this end, we introduce a novel \textit{region-dependent dual-clock denoising process}, which assigns one of two distinct SDE timesteps to different regions across frames: strong alignment for user-specified motion and weaker alignment for unconstrained areas, allowing spatially varying conditioning strength. 
To realize this effect without retraining the model, we employ a simple yet effective diffusion blending strategy akin to~\citep{avrahami2022blended,lugmayr2022repaint}.
This allows the model to adhere closely to a specified motion where it exists while allowing greater freedom to invent plausible dynamics elsewhere.

Unlike prior approaches that rely solely on (either sparse or dense) displacement fields as a guiding signal, our method is conditioned directly on the reference video itself.
This provides a richer supervisory signal: In regions where alignment is strongly enforced, we not only constrain motion, but can also dictate appearance attributes such as color, shape, or style. 
As a result, \shortname enables \emph{joint control of motion and appearance}, extending the conditioning space beyond motion-only interfaces.
We exploit this capability to support appearance-sensitive prompting in tandem with motion control, for example, animating an object along a user-specified trajectory while simultaneously changing its color (See Fig.~\ref{Fig:apperence_control}).
In summary, our work makes the following contributions:
\begin{itemize}[leftmargin=*]

\item \textbf{Training-free motion control using crude animations.} We show that simple user-provided animations (e.g., cut-and-drag manipulations or depth-based reprojections) serve as effective motion proxies.
Adapting SDEdit-style noise injection to video diffusion and anchoring appearance with image conditioning converts these coarse inputs into realistic motion without training.
\item \textbf{Region-dependent dual-clock denoising.} We propose a denoising process that operates with two distinct noise schedules—strong alignment in motion-specified regions and weaker alignment elsewhere—enabling spatially varying conditioning without retraining.
\item \textbf{Joint motion-appearance control.} Conditioning on full reference frames, rather than on motion trajectories alone, enables simultaneous control of both motion and appearance, a capability previously limited to ambiguous text prompts.
\end{itemize}

Extensive experiments show that \shortname consistently ranks among the best performing methods on both object and camera motion benchmarks, outperforming even training-based baselines. 
The approach is training-free and plug-and-play, validated across three I2V backbones, and matches the efficiency of standard video sampling.

\vspace{-4pt}
\section{Related Work}
\label{related}
\vspace{-4pt}

Video generation has recently shown great progress, achieving high visual quality and temporal coherence \citep{blattmann2023stable,hong2022cogvideo,yang2024cogvideox}, demonstrating its utility across diverse tasks \citep{Rotstein_2025_CVPR, wiedemer2025video, voleti2024sv3d}.
However, relying solely on a prompt or a single starting frame provides limited control, particularly in specifying accurate motion.

\paragraph{Learning motion control in video generation.}
A common strategy in motion control is to \emph{learn} a trajectory-conditioned representation and fuse it throughout the network. Concretely, methods inject user trajectories via multi-scale fusion of trajectory maps in U-Net blocks \citep{yin2023dragnuwa}, parameter-efficient LoRA modules that decouple camera and object motion \citep{li2025image}, or motion-patch tokens integrated across transformer blocks \citep{zhang2025tora}; ATI encodes point tracks as Gaussian-weighted latent features \citep{wang2025ati}, TrackGo inserts auxiliary branches into SVD’s temporal self-attention \citep{zhou2025trackgo,blattmann2023stable}, and MotionPro uses region-wise trajectories plus a motion mask to distinguish object vs.\ camera motion \citep{zhang2025motionpro}.
Other methods, like ours, incorporate explicit motion-based cues rather than relying solely on learned injection.
DragAnything extracts entity representations from first-frame diffusion features and injects trajectory conditioning via a ControlNet-style branch \citep{zhang2023adding}.
Motion Prompting conditions a video diffusion model on point-track “motion prompts” via a track-conditioned ControlNet \citep{geng2025motion}.
For camera control, GEN3C \citep{Ren_2025_CVPR} constructs a depth-lifted 3D cache and, similar to our approach, conditions generation on projected renders along the target camera path.
Go-with-the-Flow \citep{burgert2025go} warps diffusion noise to align with the intended motion. While this shares our architecture-agnostic goal, unlike our method and the approaches above, it extensively fine-tunes the base model to leverage the input conditioning.

\paragraph{Training-free Motion-controllable video generation.}
Several approaches avoid additional training by reusing pretrained models.
Recent \emph{text-to-video} (T2V) approaches manipulate attention to control motion: TrailBlazer modifies spatial and temporal attention early in denoising \citep{ma2024trailblazer}, PEEKABOO gates regions using masked spatio-temporal attention \citep{jain2024peekaboo}, and FreeTraj shapes low-frequency noise with attention biases to follow bounding boxes \citep{qiu2024freetraj}.
However, these models tie bounding boxes to text, limiting part-level motion and preventing precise appearance control or in-place animation of a given image.
In parallel, \emph{motion transfer} methods synthesize videos by applying motion from a driving sequence to a still image, yet depend on a suitable reference video that is difficult to obtain \citep{jeong2024vmc,yatim2024space,pondaven2025video}.
Targeting \emph{I2V} without reference videos, SG-I2V \citep{namekata2024sg} enforces cross-frame consistency by replacing each frame’s spatial self-attention keys/values with those of the first, then optimizes the latent with a box-restricted similarity loss, and re-injects high-frequency detail via an FFT.
Although zero-shot and conceptually aligned with our concept of aligning moved objects to their first-frame representation,
This method is demonstrated on SVD-specific layers, so generality for other backbones is unclear; moreover, as shown in Sec. \ref{sec:obj-motion}, it often induces unintended camera motion.
Finally, \citep{yu2024zero} proposes a training-free trajectory-guided I2V using gated self-attention for layout-conditioned control, temporal attention for propagation, and a Motion Afterimage Suppression step. 
Its modular inpainting design inherits limitations—fidelity tied to T2I inpainting and grounding tokens, heuristic handling of large displacements, while also being designed for a specific video generation model.

\paragraph{Heterogeneous Denoising.}
Asynchronous denoising has been explored in several contexts.
RAD reformulates inpainting with element-wise noise schedules and spatial timestep embedding, enabling region-asynchronous denoising while adapting a pretrained model via LoRA \citep{kim2025rad}.
SVNR addresses spatially variant sensor noise by training with per-pixel timesteps and starting the reverse process directly from the noisy input \citep{pearl2023svnr}.
Diffusion Forcing (DF) introduces temporal heterogeneity by assigning each token (e.g., a video frame) its own noise level during training, and at sampling uses a 2D scheduling matrix over time and noise levels so tokens are denoised at different rates \citep{chen2024diffusion}.
In contrast to RAD, SVNR, and DF, our approach is training-free: We impose region-specific schedules directly at inference.
RePaint \citep{lugmayr2022repaint} is also training-free, but repeatedly re-noises unmasked regions and denoises solely the mask, so only part of the image is actively denoised.
Our method heterogeneously denoises the entire image, eliminating RePaint’s resampling loops since no region is excluded.

\vspace{-4pt}
\section{Method}
\label{method}
\vspace{-4pt}


\begin{figure}[t]
  \centering
  \includegraphics[width=\linewidth]{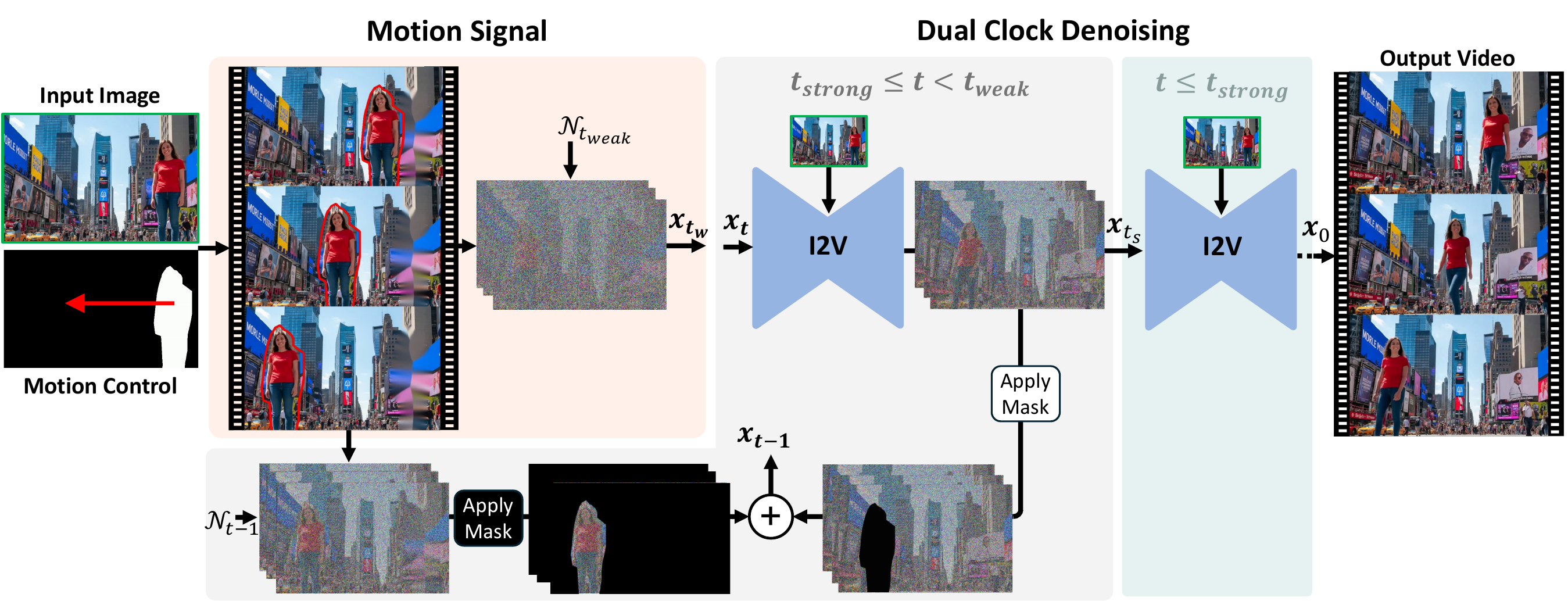}
  \vspace{-8pt}
  \caption{\textbf{Overview of \longname.} Given an input image and a motion instruction, a mask marks the region under strong control. A motion signal is then generated automatically and, together with the image, conditions an image-to-video (I2V) diffusion model. During sampling, denoising starts at different noise levels—lower inside the mask to enforce the specified motion, and higher outside to allow natural deviations in the background. Joint sampling then yields a realistic video that preserves input details while accurately following the motion control.}
  \vspace{-6pt}
  \label{fig:method}
\end{figure}


Our goal is to enable precise motion control in generative video models. Inspired by \emph{SDEdit}, which injects coarse edits into images via noising and denoising, we treat a crude warped animation (Sec.~\ref{sec:warped}) as the video analogue of such edits and adapt \emph{SDEdit} to inject intended motion into video diffusion (Sec.~\ref{sec:single_clock}).
To avoid the loss of identity that occurs when noising alone drives the process—an inherent limitation of \emph{SDEdit}—we opt for \emph{image conditioning}, anchoring the generation to the clean first frame so that the appearance is preserved throughout the sequence. 
Building on these foundations, we introduce a novel dual-clock denoising process (Sec.~\ref{sec:dual_clock}) that assigns different noise levels to distinct regions, allowing spatially varying motion guidance.
An overview of these components is shown in Fig.~\ref{fig:method}.
Finally, our procedure naturally extends to appearance control, allowing simultaneous specification of both dynamic and visual attributes.

\paragraph{Problem Formulation.}\label{sec:formulation}
Our method takes as input (i) a single image \(I \in \mathbb{R}^{3 \times H \times W}\), (ii) a coarse, user-specified warped reference video with \(F\) frames, \(V^{w} \in \mathbb{R}^{F \times 3\times H \times W}\), and (iii) a binary mask video \(M \in \{0,1\}^{F \times H \times W}\) indicating, for each frame, the regions where motion guidance is provided by the reference video.
The objective is to generate a realistic video \(x_0 \in \mathbb{R}^{F \times 3\times H \times W}\) that maintains fidelity to the input image while accurately following the prescribed motion. 

\subsection{Motion Signal}\label{sec:warped}
We begin by describing how the motion signal \(V^w\) is generated. To facilitate user-friendly interaction, the user selects a region in the first frame to produce an initial binary mask \(M_0\), then specifies a coarse motion by dragging this region along a trajectory, yielding the sequence \(M\).
This defines a piecewise-smooth displacement field within the masked region, which induces per-frame warps of the input image. The warped video \(V^w\) is obtained by forward warping \(I\), with identity mapping outside the mask. Forward warping naturally introduces disocclusions where previously occluded background becomes visible; these holes are filled using a simple nearest-neighbor inpainting strategy.  
Although presented here as dragging, both \(V^w\) and \(M\) can be constructed in multiple ways.
For example, in Sect.~\ref{sec:cam-motion} we show that \(V^w\) can also be produced by pixel-wise warping of the input image according to monocular depth estimation.
As demonstrated in our \href{https://time-to-move.github.io/}{{demo page}}, we support additional interactions, such as rotation and scaling of the selected region, which integrate seamlessly into the same formulation. 
While such warped animations are visually unrealistic, they faithfully capture the user-intended object placement and temporal structure. We exploit these properties by using them as a guiding signal for the video diffusion model during generation, and note that \(V^w\) can also encode appearance modifications, such as color changes, within the same framework.


\vspace{-3pt}
\subsection{SDEdit Adaptation for Motion Injection}\label{sec:single_clock}
\vspace{-3pt}
Inspired by the use of coarsely edited images as structural priors in \emph{SDEdit}~\citep{meng2021sdedit}, we adapt the method to videos by using the crudely warped video $V^w$ \emph{itself} as a coarse yet explicit motion instruction. 
We initialize sampling from a noisy version of the warped reference, 
\(
x_{t^*} \sim q(x_{t^*}\mid V^w),
\)
Previous publications~\citep{shaulov2025flowmo} show that coarse motion is determined early in the denoising trajectory; By noising $V^w$ to $t^*$, the intended dynamics are injected at precisely this stage. 
If we were to apply this procedure in a text-conditioned video diffusion model, the fidelity to the input image would be quickly lost: The model’s only knowledge of appearance comes from the noised $V^w$, so fine details cannot be preserved. To overcome this limitation, we instead opt for an \emph{image-conditioned} video diffusion model, which anchors generation to the clean first frame $I$. The resulting sampling process,
\(
x_0 \sim p_\theta(x_0 \mid x_{t^*}, I),
\)
faithfully integrates the motion guidance from $V^w$ while preserving identity and appearance throughout the generated sequence. 

\vspace{-3pt}
\subsection{Region-Dependent Dual-Clock Denoising}\label{sec:dual_clock}
\vspace{-3pt}

\begin{figure}[t]
  \centering
    \vspace{-4pt}
  \includegraphics[width=1.0\linewidth]{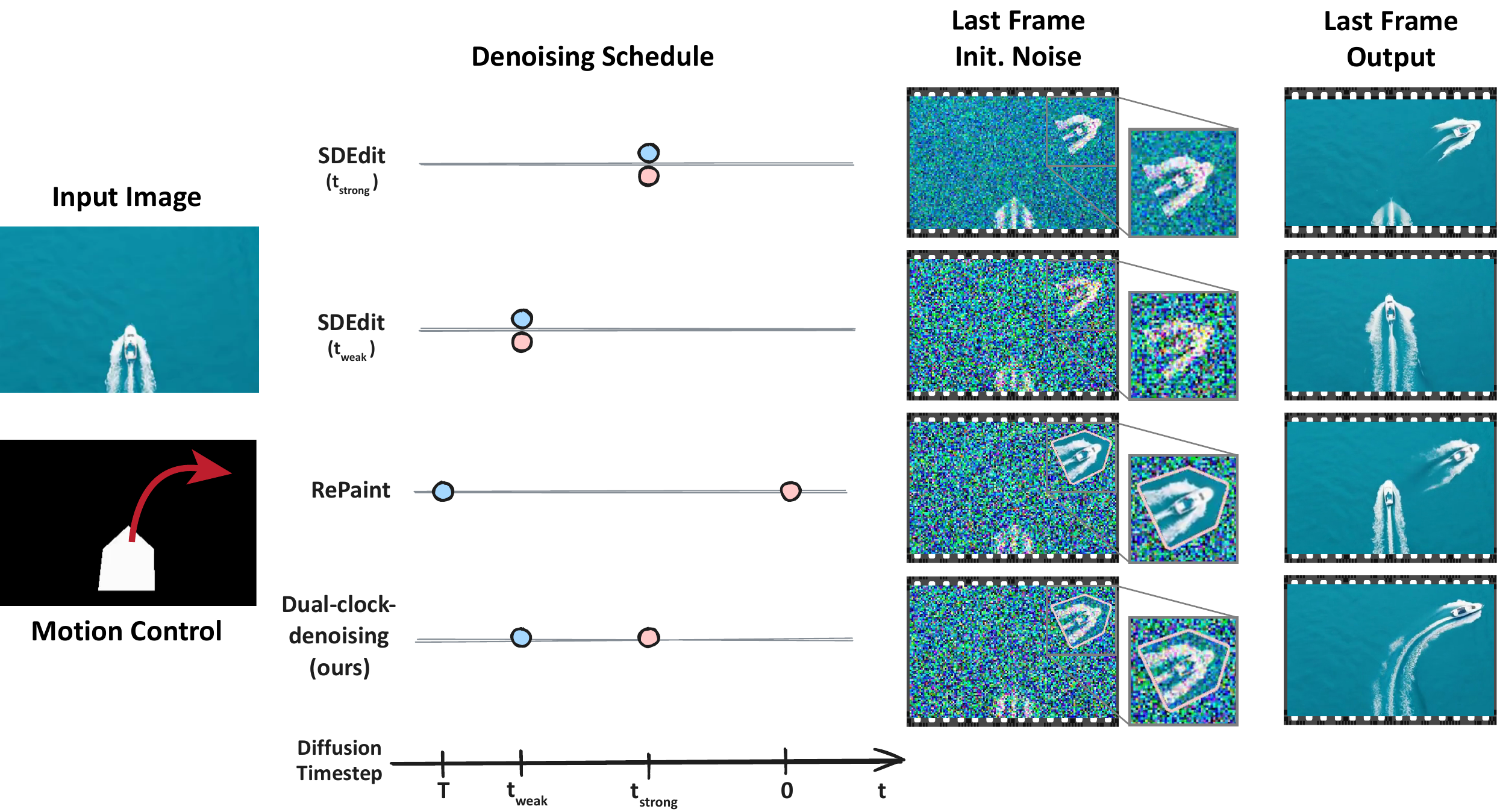}
    \vspace{-8pt}
  \caption{\textbf{Region-dependent denoising strategies.} SDEdit (single clock): low noise levels overconstrain the video, suppressing non-masked region dynamics; high noise levels improve realism but drift from the prescribed motion. RePaint (foreground override): motion is enforced in the object, but uncontrolled regions exhibit artifacts such as duplication. Dual-clock (ours): masked regions follow the intended motion with strong fidelity, while the background denoises more freely, yielding realistic dynamics without artifacts.}\label{Fig:dual_clock}
\vspace{-6pt}
\end{figure}

SDEdit employs a single noising timestep $t^*$ to corrupt the reference signal before denoising. In our setting, this creates a trade-off. The warped video $V^w$ contains regions where motion is explicitly specified (masked areas), alongside regions without explicit instruction. 
For the masked regions, we want the generated video to closely follow the prescribed motion.
In unmasked areas, we do not want them to stay static; instead, they should adapt naturally to support the motion.
For example, in Fig.~\ref{Fig:dual_clock}, when the boat is cut and dragged to follow a trajectory, the wake of the boat should modify accordingly, even though it was not directly manipulated. 
With a single timestep, SDEdit cannot accommodate this asymmetry. 
If $t^*$ is small, the denoised video adheres closely to the warped signal but inherits artifacts such as frozen backgrounds (top row). 
If $t^*$ is large, the results look realistic but drift away from the intended motion (second row).  
We therefore conjecture that different regions require different effective noising levels: masked regions demand \emph{strong adherence} to the motion signal, achieved with less noising ($t_{\text{strong}}$), while unmasked regions benefit from \emph{weaker enforcement}, achieved with increased noising ($t_{\text{weak}}$).

The challenge is that standard pretrained diffusion models, which assume inputs are corrupted by a single uniform noise level, cannot directly accommodate region-dependent noise.
To overcome this, we propose \emph{dual-clock denoising}. 
Given a mask $M$, we noise the warped video reference $V^w$ to timestep $t_{\text{weak}}$ and initialize the denoising process.
At each denoising step $t$ with $t_{\text{strong}} \leq t < t_{\text{weak}}$, we override the masked region with the corresponding region of the warped video noised to $t-1$.
This constrains the masked regions to follow the intended trajectory, while the background is free to denoise more aggressively and achieve realism.
Once $t = t_{\text{strong}}$, we stop overriding and continue the standard sampling process, allowing the model to refine both regions for a coherent result. 
%
Let $x_t$ denote the noisy sample at timestep $t$, and $\hat{x}_{t-1}(x_t
,t)$ the denoiser prediction. The update rule is 
\[
x_{t-1} \;\leftarrow\; (1 - M)\odot \hat{x}_{t-1}(x_t,t,I) \;+\; M \odot x^{w}_{t-1},
\]  
where $x^{w}_{t-1}$ is the warped reference video noised to timestep $t-1$. 


\vspace{-2pt}
\paragraph{Efficiency and Applicability.}
Our method is a lightweight modification to standard sampling, adding no extra computation or latency over regular video diffusion. 
It is entirely training-free and plug-and-play for image-conditioned I2V models; In experiments, it integrates with three backbones, demonstrating broad applicability.

\begin{figure}[t]
  \centering
  \vspace{-8pt}
  \includegraphics[width=1.0\linewidth]{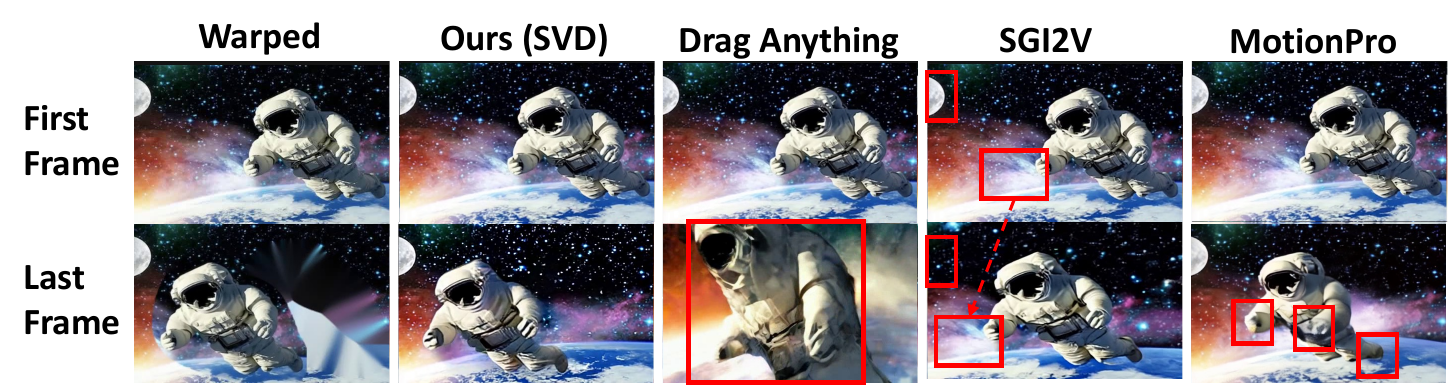}
  \vspace{-14pt}
  \caption{\textbf{Qualitative comparison on MC-Bench}
     Competing methods exhibit artifacts (red), whereas \shortname achieves clean placement and appearance consistency.
     }
  \label{fig:object_drag_banchmark}
\end{figure}

\begin{table}
    \centering
    \begin{tabular}{l|@{}>{\small}c@{}|
        @{\hspace{1pt}}>{\small}c@{} 
        >{\hspace{2pt}\small}c@{}|
        >{\hspace{-1pt}\small}c@{\hspace{3pt}}
        >{\small}c@{\hspace{3pt}}
        >{\small}c@{\hspace{3pt}}
        >{\small}c@{\hspace{3pt}}
        >{\small}c@{\hspace{3pt}}
        >{\small}c@{\hspace{3pt}}}
        \toprule
        Method & \begin{tabular}{@{}c@{}}Training\\Free?\end{tabular} &
        \begin{tabular}{@{}c@{}}CTD$_\downarrow$\end{tabular} &
        \begin{tabular}{@{}c@{}}BG--Obj\\CTD$_\uparrow$\end{tabular} &
        \begin{tabular}{@{}c@{}}\fontsize{8pt}{9pt}\selectfont{Dynamic}\\\fontsize{8pt}{9pt}\selectfont{Degree}$_\uparrow$\end{tabular} &
        \begin{tabular}{@{}c@{}}\fontsize{8pt}{9pt}\selectfont{Subject}\\\fontsize{8pt}{9pt}\selectfont{Consistency}$_\uparrow$\end{tabular} &
        \begin{tabular}{@{}c@{}}\fontsize{8pt}{9pt}\selectfont{Background}\\\fontsize{8pt}{9pt}\selectfont{Consistency}$_\uparrow$\end{tabular} &
        \begin{tabular}{@{}c@{}}\fontsize{8pt}{9pt}\selectfont{Motion}\\\fontsize{8pt}{9pt}\selectfont{Smoothness}$_\uparrow$\end{tabular} &
        \begin{tabular}{@{}c@{}}\fontsize{8pt}{9pt}\selectfont{Aesthetic}\\\fontsize{8pt}{9pt}\selectfont{Quality}$_\uparrow$\end{tabular} &
        \begin{tabular}{@{}c@{}}\fontsize{8pt}{9pt}\selectfont{Imaging}\\ \fontsize{8pt}{9pt}\selectfont{Quality}$_\uparrow$\end{tabular} \\
        \midrule
        \multicolumn{10}{c}{\textit{SVD-Based Models}} \\
        \midrule
        DragAnything & \xmark  & 10.645 & \textbf{50.885} & \textbf{0.981} & 0.956 & 0.942 & 0.983 & 0.531 & 0.554 \\
        SG-I2V$^*$ & \cmark & \textbf{5.796} & 12.042 & 0.803 & 0.976 & 0.953 & 0.991 & 0.553 & 0.621 \\
        MotionPro   & \xmark  & 8.685 & 24.485 & 0.422 & \textbf{0.979} & \textbf{0.975} & \textbf{0.993} & \textbf{0.559} & \textbf{0.617} \\
        Ours        & \textbf{\cmark} & 7.967 & 35.340 & 0.427 & \textbf{0.979} & 0.967 & \textbf{0.993} & 0.548 & \textbf{0.617} \\
        \midrule
        \multicolumn{10}{c}{\textit{CogVideoX-Based Models with Longer Generated Videos}} \\
        \midrule
        GWTF$_{\gamma=0.7}$ & \xmark & 32.548 & 86.614 & 0.736 & 0.963 & 0.965 & 0.989 & 0.517 & 0.539 \\
        GWTF$_{\gamma=0.5}$ & \xmark & 27.844 & \textbf{87.708} & \textbf{0.764} & 0.958 & 0.963 & 0.988 & 0.513 & 0.539 \\
        Ours & \cmark & \textbf{13.665} & 70.608 & 0.357 & \textbf{0.980} & \textbf{0.977} & \textbf{0.995} & \textbf{0.531} & \textbf{0.579} \\
        \bottomrule
    \end{tabular}
    \vspace{-8pt}
    \caption{\noindent\
    \textbf{Quantitative results on MC-Bench object motion control.}
    }
      \vspace{-10pt}
    \label{tbl:main_table}
\end{table}

\vspace{-4pt}
\section{Experiments}
\label{experiments}
\vspace{-4pt}
We evaluate \shortname in three complementary settings: Object motion control (Sec.~\ref{sec:obj-motion}), camera motion control (Sec.~\ref{sec:cam-motion}), and joint motion–appearance editing (Sec.~\ref{sec:app-ctrl}). These cover the primary modes of user intent: animating a selected object, inducing global motion via viewpoint changes, and modifying the appearance of scene elements. For the first two, we report quantitative benchmarks and qualitative comparisons against state-of-the-art training-based and training-free baselines. For appearance editing, where no standard benchmark exists, we present qualitative results highlighting capabilities unique to our approach. We also demonstrate plug-and-play generality across multiple I2V backbones (Sec.~\ref{sec:we_wan}) and analyze the dual-clock schedule via ablations (App.~\ref{apdx:ablation}). Video demonstrations are included in the \href{https://time-to-move.github.io/}{{project page}}. 

\vspace{-4pt}
\subsection{Object Motion Control}
\label{sec:obj-motion}
\vspace{-4pt}
We evaluate \shortname for object-level motion control. The inputs are a single source image, a binary mask of the target object, and a 2D trajectory defining the desired motion. 
We benchmark on MC-Bench~\citep{zhang2025motionpro} under its official protocol.
Further details of the evaluation protocol and implementation are provided in App.~\ref{apdx:object_control}.

\noindent\textbf{Baselines.}
We compare against both training-based methods—DragAnything~\citep{wu2024draganything}, MotionPro~\citep{zhang2025motionpro}, and Go-With-the-Flow (GWTF)~\citep{burgert2025go}—and the training-free SG-I2V~\citep{namekata2024sg}. 
Results are grouped by backbone: SVD (hybrid conv/attention, $\sim$1.5B parameters) and CogVideoX (Diffusion Transformer, 5B parameters). 
We apply our backbone-agnostic method to both architectures, denoting them as \shortnamesvd and \shortnamecog.
For fairness, we report GWTF with both recommended noise-degradation values (\(\gamma \in \{0.5, 0.7\}\)). 

\noindent\textbf{Metrics.}
We evaluate motion adherence and perceptual quality. For adherence, we use MC-Bench’s CoTracker Distance (CTD) for object trajectories and BG–Obj CTD to detect unintended background co-motion. For perceptual quality, we adopt VBench~\citep{huang2024vbench}, a reference-free suite of automated video metrics. See App.~\ref{apdx:object_control} for more details.

\begin{figure}[t]
  \centering
  \includegraphics[width=1.0\linewidth]{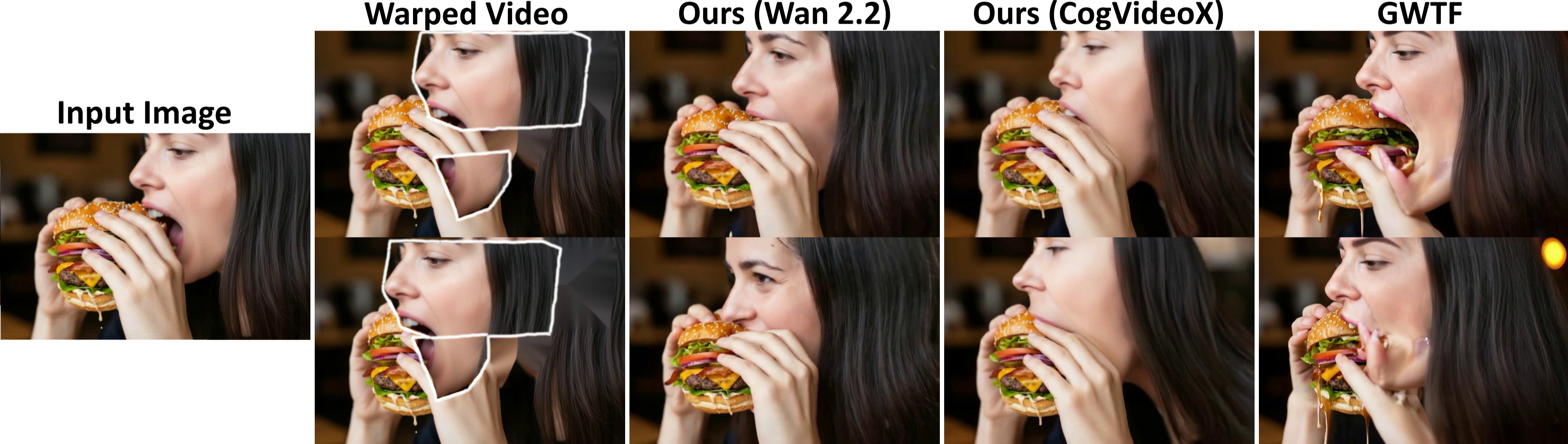}
  \vspace{-17pt}
  \caption{\textbf{Comparison on a challenging cut-and-drag example.} GWTF exhibits strong artifacts under large motion (right); \shortname follows the prescribed motion realistically across various models.}
\vspace{-2pt}
    \label{fig:object_drag_human_crafted}
\end{figure}
\begin{table}[t]
\centering
\begingroup
\renewcommand{\arraystretch}{1.08} 
\begin{tabular}{l|>{\small}c
        @{\hspace{3pt}}>{\small}c
        >{\hspace{3pt}}>{\small}c
        >{\hspace{3pt}}>{\small}c
        >{\hspace{3pt}}>{\small}c
        >{\hspace{3pt}}>{\small}c
        >{\hspace{3pt}}>{\small}c}
    \toprule
    Method &
    \begin{tabular}{@{}c@{}}  MSE  $_\downarrow$\end{tabular} &
    \begin{tabular}{@{}c@{}}FID$_\downarrow$\end{tabular} &
    \begin{tabular}{@{}c@{}}LPIPS$_\downarrow$\end{tabular} &
    \begin{tabular}{@{}c@{}}SSIM$_\uparrow$\end{tabular} &
    \begin{tabular}{@{}c@{}}CLIP\\Cons.$_\uparrow$\end{tabular} &
    \begin{tabular}{@{}c@{}}Optical\\flow$_\downarrow$\end{tabular} \\
    \midrule
    GWTF$_{\gamma=0.5}$ & 0.033 & 25.990 & 0.371 & 0.526 & 0.981 & 76.714 \\
    GWTF$_{\gamma=0.7}$ & 0.042 & 28.483 & 0.370 & 0.410 & \textbf{0.985} & 81.738 \\
    Warped  & 0.025 & 33.443 & 0.339 & 0.560 & 0.981 & 65.494 \\
    Ours    & \textbf{0.022} & \textbf{21.966} & \textbf{0.332} & \textbf{0.586} & 0.983 & \textbf{60.558} \\
    \bottomrule
\end{tabular}
\endgroup
  \vspace{-6pt}
    \caption{\noindent\textbf{Quantitative results on DL3DV camera motion control.}
    }
  \vspace{-10pt}
\label{tbl:dl3dv}
\end{table}

\noindent\textbf{Results.}
Tab.~\ref{tbl:main_table} summarizes the results. 
Across both backbones, \shortname attains the lowest CoTracker distance (best adherence to the prescribed motion), excluding SG-I2V. 
On SVD, our VBench quality matches MotionPro, with minor metric trade-offs, and surpasses DragAnything and SG-I2V on most measures.
\shortname's dynamic degree is lower than DragAnything and SG-I2V: we attribute this to DragAnything often inducing unintended scene motion and local deformations (Fig.~\ref{fig:object_drag_banchmark}), whereas SG-I2V frequently triggers camera co-motion, moving the whole scene rather than just the object (e.g., a rightward pan in the same figure, where the camera shifts right and the moon exits the frame).
This effect—also noted by \citet{burgert2025go}—is reflected in SG-I2V’s substantially lower \mbox{BG--Obj CTD}, indicating strong object–background co-motion.
On the CogVideoX backbone, \shortname achieves substantially stronger adherence to motion conditioning and higher scores on nearly all video-quality metrics compared to GWTF. 
The only exception is the “dynamic” score, where GWTF reports higher values; however, these gains often come at the cost of scene deformations and inconsistencies, as evident from the background- and subject-consistency metrics in Tab.~\ref{tbl:main_table} and in Fig.~\ref{fig:object_drag_human_crafted}.
Overall, \shortname exceeds the performance of both training-based and training-free baselines on most metrics, while remaining entirely training-free.

\vspace{-2pt}
\noindent\textbf{Qualitative Examples.}
In Fig.~\ref{fig:object_drag_banchmark}, we present a representative example from the MC-Bench benchmark, using SVD as the common I2V backbone. Competing methods introduce noticeable artifacts (highlighted in red), while our \shortname produces clean foreground placement at the intended location and preserves fidelity to the first-frame appearance.
Additional videos and benchmark results are provided in Fig.~\ref{fig:object_drag_human_crafted}, App.~\ref{apdx:qual_comp}, and on our \href{https://time-to-move.github.io/supplementary.html}{{supplementary results page}}.

\vspace{-4pt}
\subsection{Camera Motion Control}
\label{sec:cam-motion}
\vspace{-4pt}
We evaluate \shortname{} on synthesizing realistic videos from a single image under prescribed camera motion. Following GWTF, we use a subset of DL3DV-10K~\citep{ling2024dl3dv}, which contains static-scene videos with per-frame camera annotations. From the first frame, we estimate metric depth with DepthPro~\citep{Bochkovskii2024:arxiv}, back-project to a 3D point cloud, and reproject along the prescribed motion to construct a reference video. Missing regions are inpainted by nearest-neighbor color assignment, and valid pixels define a guidance mask. We evaluate 150 sequences with 49 target views each, comparing generated results against the original frames at the same viewpoints. Further details appear in App.~\ref{apdx:impl_details}.

\noindent\textbf{Baselines.}
We benchmark \shortname against GWTF, the leading prior method for free-form motion control.
The protocol for constructing depth-based warped videos is identical to that used in our approach; however,
%
GWTF further extracts optical flow from them to synthesize noise warping. 


\noindent\textbf{Metrics.}
\label{sec:dl3dv-metrics}
With ground-truth videos available, we evaluate frame-level alignment using MSE, LPIPS~\citep{Zhang18LPIPS}, and SSIM~\citep{wang2004image}. Motion consistency is assessed by the MSE between RAFT-estimated optical flows~\citep{teed2020raft} of generated and ground-truth videos. Distributional similarity is assessed with FID~\citep{Heusel17FID} between all generated and original frames. Temporal consistency is measured as the average CLIP~\citep{radford2021learning} cosine similarity between consecutive generated frames. 




\begin{figure}[t]
  \centering
  \includegraphics[width=1.0\linewidth]{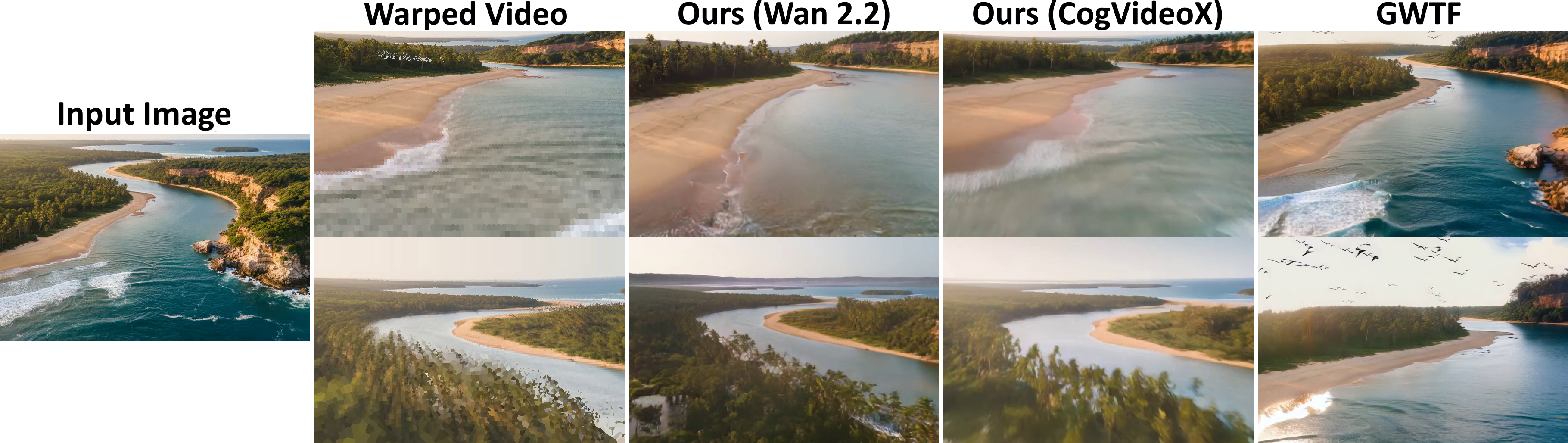}
  \vspace{-16pt}
  \caption{\textbf{Qualitative comparison of camera-motion control.} 
  GWTF drifts from the target camera path, while \shortname leverages the warped reference to enforce motion, yielding smooth, artifact-free results beyond simple depth warping.}
  \label{fig:camera_control_human_crafted}
  \vspace{-12pt}
\end{figure}

\noindent\textbf{Results.}
Tab.~\ref{tbl:dl3dv} reports quantitative results, and App.~\ref{apdx:qual_cam_ctrl} together with our \href{https://time-to-move.github.io/supplementary.html}{{video
 appendix}} present qualitative results.
Our method delivers the best camera-motion control, outperforming baselines in motion fidelity and pixel quality: vs. the best GWTF variant, pixel MSE drops by \(\mathbf{33\%}\) and FID by \(\mathbf{15.5\%}\); optical-flow MSE also decreases, indicating better temporal alignment across the scene.

\noindent\textbf{Qualitative Examples.}
Fig.~\ref{fig:camera_control_human_crafted} compares \shortname with GWTF on input images and user-specified camera trajectories.  
GWTF struggles with long motions as it relies on noise warping for scene consistency and drifts from the prescribed path. In contrast, \shortname precisely follows the target camera motion and preserves identity across frames, yielding smooth, realistic sequences. Depth warping is shown as coarse reference; \shortname removes its tearing and holes while retaining the intended motion.

\vspace{-3pt}
\subsection{Appearance Control}
\label{sec:app-ctrl}
\begin{figure}[t]
  \centering
  \includegraphics[width=1.0\linewidth]{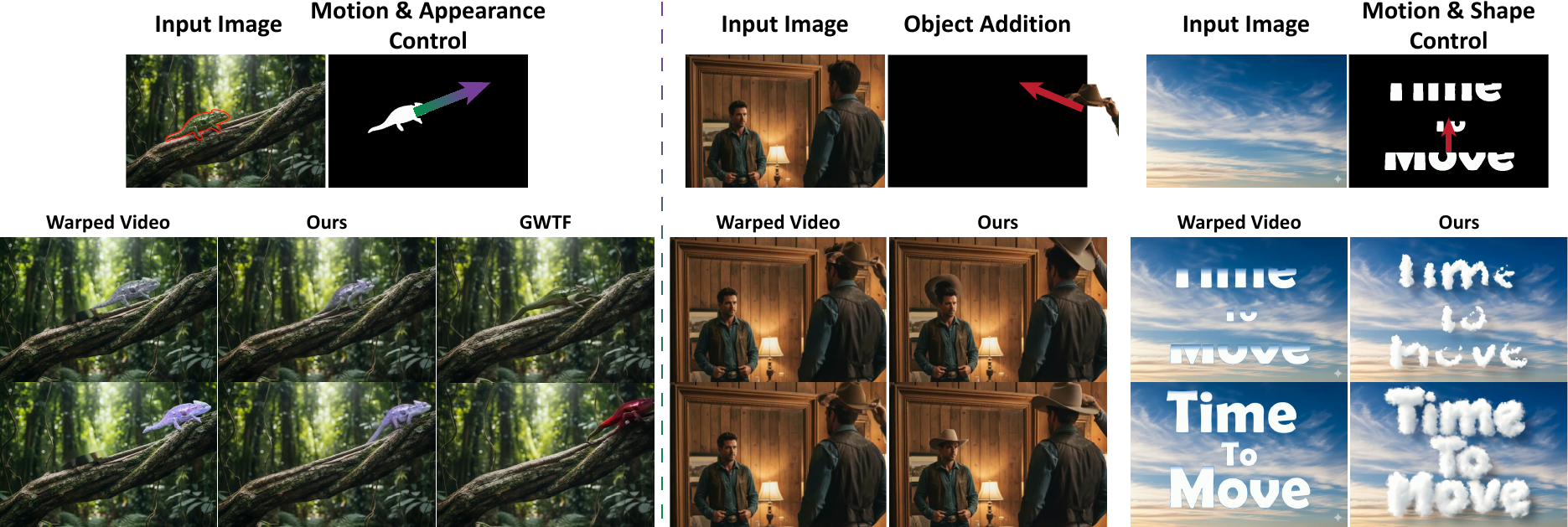}
    \vspace{-14pt}
  \caption{\textbf{Joint motion and appearance control.} 
    \shortname leverages a user-provided warped reference to control both motion and per-pixel appearance in diverse tasks, with per-task details given in \ref{sec:app-ctrl}.
    }
  \label{Fig:apperence_control}
  \vspace{-10pt}
\end{figure}

\vspace{-3pt}
Beyond motion, \shortname enables pixel-level appearance specification across the scene. 
By conditioning on full reference frames, the crude animation constrains both \emph{where} objects move and \emph{how} they look. 
Users can guide motion and evolving appearance jointly, without retraining or additional cost. In contrast, prior methods rely on trajectories and text alone, limiting them to ambiguous appearance changes. 
In Fig.~\ref{Fig:apperence_control} we illustrate three setups: 
(i) \emph{Motion and appearance control}: a chameleon follows a user-drawn trajectory while changing its color from green to purple. For comparison, GWTF is run with optical flow and a text prompt describing the desired color (see App.~\ref{app:app_control}); our method preserves both motion and appearance, whereas GWTF fails to satisfy both constraints. 
(ii) \emph{Object insertion}: conditioning on full frames allows adding new objects. We place a hat on a cowboy looking in a mirror; the hat blends naturally into the scene and appears consistently in the reflection. 
(iii) \emph{Joint motion and shape control}: \shortname preserves the intended graphic deformations while harmonizing appearance with the scene as clouds are revealed.

\vspace{-3pt}
\subsection{Plug-and-Play Model Adaptation}\label{sec:we_wan}
\vspace{-2pt}
With video generators evolving rapidly and parameter counts rising, adding motion control \emph{without retraining} becomes especially valuable.
Beyond SVD and CogVideoX, \shortname applies \emph{as is} to any image-to-video diffusion model. 
We demonstrate this on the recently released 14B-parameters WAN~2.2\footnote{\url{https://github.com/Wan-Video/Wan2.2?tab=readme-ov-file}}~\citep{wan2025wan}: with only a brief adaptation, \shortname enables both local object control and explicit camera-motion conditioning.
In Figs.~\ref{fig:teaser}, \ref{fig:object_drag_human_crafted}, and~\ref{fig:camera_control_human_crafted}, we present a set of challenging examples.
By contrast, GWTF achieves motion control only after fine-tuning CogVideoX-5B with warped-noise training, requiring \(\sim\!7{,}680\) A100-80GB GPU-hours.


\vspace{-8pt}
\section{Conclusions, Limitations and Future Directions
}
\label{sec:conclusions}
\vspace{-8pt}
We introduced a training-free framework for motion and appearance control in I2V diffusion models. 
By extending the SDEdit principle to videos, we treat warped reference animations as direct motion guidance, while image conditioning preserves fidelity to the input. 
To balance strict adherence in user-specified regions while enabling natural adaptation in the remaining regions, we proposed region-dependent dual-clock denoising, a plug-and-play strategy that produces realistic and faithful generations.  Our method has several limitations. Although our framework adapts seamlessly to different I2V backbones the dual-clock scheme still requires tuning of $(t_{\text{weak}}, t_{\text{strong}})$. 
Identity preservation is restricted to content visible in the first frame; objects entering later cannot be anchored beyond what is implicitly recovered during denoising.
Finally, our framework requires full object masks when specifying motion, unlike some motion-prompting methods that are explicitly trained to operate from partial markings.
Nevertheless, our method remains robust to imperfect masks, as demonstrated in MC-Bench.
Our framework accommodates extensions beyond our current implementation. 
In particular, the dual-clock scheme could be generalized to support multiple regions, soft masks, or smoother noise schedules, offering more fine-grained control. We leave richer appearance edits (e.g., stylistic transformations), articulated motion, and long-horizon video generation for future work.

\bibliography{iclr2026_conference}

@article{blattmann2023stable,
  title={Stable video diffusion: Scaling latent video diffusion models to large datasets},
  author={Blattmann, Andreas and Dockhorn, Tim and Kulal, Sumith and Mendelevitch, Daniel and Kilian, Maciej and Lorenz, Dominik and Levi, Yam and English, Zion and Voleti, Vikram and Letts, Adam and others},
  journal={arXiv preprint arXiv:2311.15127},
  year={2023}
}

@inproceedings{hong2022cogvideo,
  title={Cogvideo: Large-scale pretraining for text-to-video generation via transformers},
  author={Hong, Wenyi and Ding, Ming and Zheng, Wendi and Liu, Xinghan and Tang, Jie},
    booktitle={International Conference on Learning Representations},
  year={2023}
}

@inproceedings{yang2024cogvideox,
  title={Cogvideox: Text-to-video diffusion models with an expert transformer},
  author={Yang, Zhuoyi and Teng, Jiayan and Zheng, Wendi and Ding, Ming and Huang, Shiyu and Xu, Jiazheng and Yang, Yuanming and Hong, Wenyi and Zhang, Xiaohan and Feng, Guanyu and others},
  booktitle={International Conference on Learning Representations},
  year={2025}
}

@article{yin2023dragnuwa,
  title={Dragnuwa: Fine-grained control in video generation by integrating text, image, and trajectory},
  author={Yin, Shengming and Wu, Chenfei and Liang, Jian and Shi, Jie and Li, Houqiang and Ming, Gong and Duan, Nan},
  journal={arXiv preprint arXiv:2308.08089},
  year={2023}
}

@inproceedings{li2025image,
  title={Image conductor: Precision control for interactive video synthesis},
  author={Li, Yaowei and Wang, Xintao and Zhang, Zhaoyang and Wang, Zhouxia and Yuan, Ziyang and Xie, Liangbin and Shan, Ying and Zou, Yuexian},
  booktitle={Proceedings of the AAAI Conference on Artificial Intelligence},
  volume={39},  
  pages={5031--5038},
  year={2025}
}

@inproceedings{wu2024draganything,
  title={Draganything: Motion control for anything using entity representation},
  author={Wu, Weijia and Li, Zhuang and Gu, Yuchao and Zhao, Rui and He, Yefei and Zhang, David Junhao and Shou, Mike Zheng and Li, Yan and Gao, Tingting and Zhang, Di},
  booktitle={European Conference on Computer Vision},
  pages={331--348},
  year={2024},
  organization={Springer}
}

@inproceedings{zhou2025trackgo,
  title={Trackgo: A flexible and efficient method for controllable video generation},
  author={Zhou, Haitao and Wang, Chuang and Nie, Rui and Liu, Jinlin and Yu, Dongdong and Yu, Qian and Wang, Changhu},
  booktitle={Proceedings of the AAAI Conference on Artificial Intelligence},
  volume={39},
  number={10},
  pages={10743--10751},
  year={2025}
}

@inproceedings{zhang2025motionpro,
  title={MotionPro: A Precise Motion Controller for Image-to-Video Generation},
  author={Zhang, Zhongwei and Long, Fuchen and Qiu, Zhaofan and Pan, Yingwei and Liu, Wu and Yao, Ting and Mei, Tao},
  booktitle={Proceedings of the Computer Vision and Pattern Recognition Conference},
  pages={27957--27967},
  year={2025}
}

@inproceedings{zhang2025tora,
  title={Tora: Trajectory-oriented diffusion transformer for video generation},
  author={Zhang, Zhenghao and Liao, Junchao and Li, Menghao and Dai, Zuozhuo and Qiu, Bingxue and Zhu, Siyu and Qin, Long and Wang, Weizhi},
  booktitle={Proceedings of the Computer Vision and Pattern Recognition Conference},
  pages={2063--2073},
  year={2025}
}

@inproceedings{geng2025motion,
  title={Motion prompting: Controlling video generation with motion trajectories},
  author={Geng, Daniel and Herrmann, Charles and Hur, Junhwa and Cole, Forrester and Zhang, Serena and Pfaff, Tobias and Lopez-Guevara, Tatiana and Aytar, Yusuf and Rubinstein, Michael and Sun, Chen and others},
  booktitle={Proceedings of the Computer Vision and Pattern Recognition Conference},
  pages={1--12},
  year={2025}
}

@inproceedings{zhang2023adding,
  title={Adding conditional control to text-to-image diffusion models},
  author={Zhang, Lvmin and Rao, Anyi and Agrawala, Maneesh},
  booktitle={Proceedings of the IEEE/CVF international conference on computer vision},
  pages={3836--3847},
  year={2023}
}

@article{meng2021sdedit,
  title={Sdedit: Guided image synthesis and editing with stochastic differential equations},
  author={Meng, Chenlin and He, Yutong and Song, Yang and Song, Jiaming and Wu, Jiajun and Zhu, Jun-Yan and Ermon, Stefano},
  journal={arXiv preprint arXiv:2108.01073},
  year={2021}
}

@article{wang2025ati,
  title={ATI: Any Trajectory Instruction for Controllable Video Generation},
  author={Wang, Angtian and Huang, Haibin and Fang, Jacob Zhiyuan and Yang, Yiding and Ma, Chongyang},
  journal={arXiv preprint arXiv:2505.22944},
  year={2025}
}

@inproceedings{burgert2025go,
  title={Go-with-the-flow: Motion-controllable video diffusion models using real-time warped noise},
  author={Burgert, Ryan and Xu, Yuancheng and Xian, Wenqi and Pilarski, Oliver and Clausen, Pascal and He, Mingming and Ma, Li and Deng, Yitong and Li, Lingxiao and Mousavi, Mohsen and others},
  booktitle={Proceedings of the Computer Vision and Pattern Recognition Conference},
  pages={13--23},
  year={2025}
}

@inproceedings{ling2024dl3dv,
  title={Dl3dv-10k: A large-scale scene dataset for deep learning-based 3d vision},
  author={Ling, Lu and Sheng, Yichen and Tu, Zhi and Zhao, Wentian and Xin, Cheng and Wan, Kun and Yu, Lantao and Guo, Qianyu and Yu, Zixun and Lu, Yawen and others},
  booktitle={Proceedings of the IEEE/CVF Conference on Computer Vision and Pattern Recognition},
  pages={22160--22169},
  year={2024}
}

@inproceedings{Bochkovskii2024:arxiv,
  author     = {Aleksei Bochkovskii and Ama\"{e}l Delaunoy and Hugo Germain and Marcel Santos and
               Yichao Zhou and Stephan R. Richter and Vladlen Koltun},
  title      = {Depth Pro: Sharp Monocular Metric Depth in Less Than a Second},
  booktitle  = {International Conference on Learning Representations},
  year       = {2025},
  url        = {https://arxiv.org/abs/2410.02073},
}

@misc{openai2024gpt4ocard,
      title={GPT-4o System Card}, 
      author={OpenAI},
      year={2024},
      eprint={2410.21276},
      archivePrefix={arXiv},
      primaryClass={cs.CL},
      url={https://arxiv.org/abs/2410.21276}, 
}

@INPROCEEDINGS {Zhang18LPIPS,
author = { Zhang, Richard and Isola, Phillip and Efros, Alexei A. and Shechtman, Eli and Wang, Oliver },
booktitle = { 2018 IEEE/CVF Conference on Computer Vision and Pattern Recognition (CVPR) },
title = {{ The Unreasonable Effectiveness of Deep Features as a Perceptual Metric }},
year = {2018},
pages = {586-595},
doi = {10.1109/CVPR.2018.00068},
url = {https://doi.ieeecomputersociety.org/10.1109/CVPR.2018.00068},
publisher = {IEEE Computer Society},
address = {Los Alamitos, CA, USA},
month =Jun}

@inproceedings{Heusel17FID,
author = {Heusel, Martin and Ramsauer, Hubert and Unterthiner, Thomas and Nessler, Bernhard and Hochreiter, Sepp},
title = {GANs trained by a two time-scale update rule converge to a local nash equilibrium},
year = {2017},
isbn = {9781510860964},
publisher = {Curran Associates Inc.},
address = {Red Hook, NY, USA},
booktitle = {Proceedings of the 31st International Conference on Neural Information Processing Systems},
pages = {6629–6640},
numpages = {12},
location = {Long Beach, California, USA},
series = {NIPS'17}
}

@inproceedings{radford2021learning,
  title={Learning transferable visual models from natural language supervision},
  author={Radford, Alec and Kim, Jong Wook and Hallacy, Chris and Ramesh, Aditya and Goh, Gabriel and Agarwal, Sandhini and Sastry, Girish and Askell, Amanda and Mishkin, Pamela and Clark, Jack and others},
  booktitle={International conference on machine learning},
  pages={8748--8763},
  year={2021},
  organization={PmLR}
}

@inproceedings{teed2020raft,
  title={Raft: Recurrent all-pairs field transforms for optical flow},
  author={Teed, Zachary and Deng, Jia},
  booktitle={European conference on computer vision},
  pages={402--419},
  year={2020},
  organization={Springer}
}

@inproceedings{namekata2024sg,
  title={SG-I2V: Self-Guided Trajectory Control in Image-to-Video Generation},
  author={Namekata, Koichi and Bahmani, Sherwin and Wu, Ziyi and Kant, Yash and Gilitschenski, Igor and Lindell, David B},
  booktitle={International Conference on Learning Representations},
year={2025}
}

@inproceedings{jain2024peekaboo,
  title={Peekaboo: Interactive video generation via masked-diffusion},
  author={Jain, Yash and Nasery, Anshul and Vineet, Vibhav and Behl, Harkirat},
  booktitle={Proceedings of the IEEE/CVF Conference on Computer Vision and Pattern Recognition},
  pages={8079--8088},
  year={2024}
}

@inproceedings{ma2024trailblazer,
  title={Trailblazer: Trajectory control for diffusion-based video generation},
  author={Ma, Wan-Duo Kurt and Lewis, John P and Kleijn, W Bastiaan},
  booktitle={SIGGRAPH Asia 2024 Conference Papers},
  pages={1--11},
  year={2024}
}

@inproceedings{kim2025rad,
  title={Rad: Region-aware diffusion models for image inpainting},
  author={Kim, Sora and Suh, Sungho and Lee, Minsik},
  booktitle={Proceedings of the Computer Vision and Pattern Recognition Conference},
  pages={2439--2448},
  year={2025}
}

@inproceedings{avrahami2022blended,
  title={Blended diffusion for text-driven editing of natural images},
  author={Avrahami, Omri and Lischinski, Dani and Fried, Ohad},
  booktitle={Proceedings of the IEEE/CVF conference on computer vision and pattern recognition},
  pages={18208--18218},
  year={2022}
}

@inproceedings{lugmayr2022repaint,
  title={Repaint: Inpainting using denoising diffusion probabilistic models},
  author={Lugmayr, Andreas and Danelljan, Martin and Romero, Andres and Yu, Fisher and Timofte, Radu and Van Gool, Luc},
  booktitle={Proceedings of the IEEE/CVF conference on computer vision and pattern recognition},
  pages={11461--11471},
  year={2022}
}

@inproceedings{yu2024zero,
  title={Zero-shot controllable image-to-video animation via motion decomposition},
  author={Yu, Shoubin and Fang, Jacob Zhiyuan and Zheng, Jian and Sigurdsson, Gunnar and Ordonez, Vicente and Piramuthu, Robinson and Bansal, Mohit},
  booktitle={Proceedings of the 32nd ACM International Conference on Multimedia},
  pages={3332--3341},
  year={2024}
}

@article{qiu2024freetraj,
  title={Freetraj: Tuning-free trajectory control in video diffusion models},
  author={Qiu, Haonan and Chen, Zhaoxi and Wang, Zhouxia and He, Yingqing and Xia, Menghan and Liu, Ziwei},
  journal={arXiv preprint arXiv:2406.16863},
  year={2024}
}

@inproceedings{karaev2024cotracker,
  title={Cotracker: It is better to track together},
  author={Karaev, Nikita and Rocco, Ignacio and Graham, Benjamin and Neverova, Natalia and Vedaldi, Andrea and Rupprecht, Christian},
  booktitle={European conference on computer vision},
  pages={18--35},
  year={2024},
  organization={Springer}
}

@article{sand2008particle,
  title={Particle video: Long-range motion estimation using point trajectories},
  author={Sand, Peter and Teller, Seth},
  journal={International journal of computer vision},
  volume={80},
  number={1},
  pages={72--91},
  year={2008},
  publisher={Springer}
}

@inproceedings{tumanyan2024dino,
  title={Dino-tracker: Taming dino for self-supervised point tracking in a single video},
  author={Tumanyan, Narek and Singer, Assaf and Bagon, Shai and Dekel, Tali},
  booktitle={European Conference on Computer Vision},
  pages={367--385},
  year={2024},
  organization={Springer}
}

@article{pearl2023svnr,
  title={Svnr: Spatially-variant noise removal with denoising diffusion},
  author={Pearl, Naama and Brodsky, Yaron and Berman, Dana and Zomet, Assaf and Acha, Alex Rav and Cohen-Or, Daniel and Lischinski, Dani},
  journal={arXiv preprint arXiv:2306.16052},
  year={2023}
}

@article{chen2024diffusion,
  title={Diffusion forcing: Next-token prediction meets full-sequence diffusion},
  author={Chen, Boyuan and Mart{\'\i} Mons{\'o}, Diego and Du, Yilun and Simchowitz, Max and Tedrake, Russ and Sitzmann, Vincent},
  journal={Advances in Neural Information Processing Systems},
  volume={37},
  pages={24081--24125},
  year={2024}
}

@inproceedings{jeong2024vmc,
  title={Vmc: Video motion customization using temporal attention adaption for text-to-video diffusion models},
  author={Jeong, Hyeonho and Park, Geon Yeong and Ye, Jong Chul},
  booktitle={Proceedings of the IEEE/CVF Conference on Computer Vision and Pattern Recognition},
  pages={9212--9221},
  year={2024}
}

@inproceedings{pondaven2025video,
  title={Video motion transfer with diffusion transformers},
  author={Pondaven, Alexander and Siarohin, Aliaksandr and Tulyakov, Sergey and Torr, Philip and Pizzati, Fabio},
  booktitle={Proceedings of the Computer Vision and Pattern Recognition Conference},
  pages={22911--22921},
  year={2025}
}

@inproceedings{yatim2024space,
  title={Space-time diffusion features for zero-shot text-driven motion transfer},
  author={Yatim, Danah and Fridman, Rafail and Bar-Tal, Omer and Kasten, Yoni and Dekel, Tali},
  booktitle={Proceedings of the IEEE/CVF Conference on Computer Vision and Pattern Recognition},
  pages={8466--8476},
  year={2024}
}

@inproceedings{huang2024vbench,
  title={Vbench: Comprehensive benchmark suite for video generative models},
  author={Huang, Ziqi and He, Yinan and Yu, Jiashuo and Zhang, Fan and Si, Chenyang and Jiang, Yuming and Zhang, Yuanhan and Wu, Tianxing and Jin, Qingyang and Chanpaisit, Nattapol and others},
  booktitle={Proceedings of the IEEE/CVF Conference on Computer Vision and Pattern Recognition},
  pages={21807--21818},
  year={2024}
}

@article{wang2004image,
  title={Image quality assessment: from error visibility to structural similarity},
  author={Wang, Zhou and Bovik, Alan C and Sheikh, Hamid R and Simoncelli, Eero P},
  journal={IEEE transactions on image processing},
  volume={13},
  number={4},
  pages={600--612},
  year={2004},
  publisher={IEEE}
}

@article{wan2025wan,
  title={Wan: Open and advanced large-scale video generative models},
  author={Wan, Team and Wang, Ang and Ai, Baole and Wen, Bin and Mao, Chaojie and Xie, Chen-Wei and Chen, Di and Yu, Feiwu and Zhao, Haiming and Yang, Jianxiao and others},
  journal={arXiv preprint arXiv:2503.20314},
  year={2025}
}

@article{shaulov2025flowmo,
  title={FlowMo: Variance-Based Flow Guidance for Coherent Motion in Video Generation},
  author={Shaulov, Ariel and Hazan, Itay and Wolf, Lior and Chefer, Hila},
  journal={arXiv preprint arXiv:2506.01144},
  year={2025}
}

@article{comanici2025gemini,
  title={Gemini 2.5: Pushing the frontier with advanced reasoning, multimodality, long context, and next generation agentic capabilities},
  author={Comanici, Gheorghe and Bieber, Eric and Schaekermann, Mike and Pasupat, Ice and Sachdeva, Noveen and Dhillon, Inderjit and Blistein, Marcel and Ram, Ori and Zhang, Dan and Rosen, Evan and others},
  journal={arXiv preprint arXiv:2507.06261},
  year={2025}
}

@InProceedings{Rotstein_2025_CVPR,
    author    = {Rotstein, Noam and Yona, Gal and Silver, Daniel and Velich, Roy and Bensaid, David and Kimmel, Ron},
    title     = {Pathways on the Image Manifold: Image Editing via Video Generation},
    booktitle = {Proceedings of the IEEE/CVF Conference on Computer Vision and Pattern Recognition (CVPR)},
    month     = {June},
    year      = {2025},
    pages     = {7857-7866}
}

@article{wiedemer2025video,
  title={Video models are zero-shot learners and reasoners},
  author={Wiedemer, Thadd{\"a}us and Li, Yuxuan and Vicol, Paul and Gu, Shixiang Shane and Matarese, Nick and Swersky, Kevin and Kim, Been and Jaini, Priyank and Geirhos, Robert},
  journal={arXiv preprint arXiv:2509.20328},
  year={2025}
}

@inproceedings{voleti2024sv3d,
  title={Sv3d: Novel multi-view synthesis and 3d generation from a single image using latent video diffusion},
  author={Voleti, Vikram and Yao, Chun-Han and Boss, Mark and Letts, Adam and Pankratz, David and Tochilkin, Dmitry and Laforte, Christian and Rombach, Robin and Jampani, Varun},
  booktitle={European Conference on Computer Vision},
  pages={439--457},
  year={2024},
  organization={Springer}
}

@InProceedings{Ren_2025_CVPR,
    author    = {Ren, Xuanchi and Shen, Tianchang and Huang, Jiahui and Ling, Huan and Lu, Yifan and Nimier-David, Merlin and M\"uller, Thomas and Keller, Alexander and Fidler, Sanja and Gao, Jun},
    title     = {GEN3C: 3D-Informed World-Consistent Video Generation with Precise Camera Control},
    booktitle = {Proceedings of the IEEE/CVF Conference on Computer Vision and Pattern Recognition (CVPR)},
    month     = {June},
    year      = {2025},
    pages     = {6121-6132}
}
\bibliographystyle{iclr2026_conference}
\appendix
\newpage

\section{Ablation Study: Dual-Clock Denoising}\label{apdx:ablation}

\begin{wraptable}{r}{0.39\linewidth} 
\vspace{-\baselineskip}              
\centering
    \begin{tabular}{@{}%
        >{\small}c@{}%
        @{\hspace{4pt}}>{\small}c@{}|%
        @{\hspace{4pt}}>{\small}c@{\hspace{4pt}}%
        >{\small}c@{\hspace{2pt}}%
        >{\small}c@{\hspace{2pt}}%
        }
        \toprule
        \begin{tabular}{@{}c@{}}First\\tick\\($t_1$)\end{tabular} &
        \begin{tabular}{@{}c@{}}Second\\tick\\($t_2$)\end{tabular} &
        \begin{tabular}{@{}c@{}}CoTracker\\distance\end{tabular} &
        \begin{tabular}{@{}c@{}}Dynamic\\degree\end{tabular} &
        \begin{tabular}{@{}c@{}}Imaging\\quality\end{tabular}
        \\
        \midrule
        \(t_{\text{weak}}\) & \(t_{\text{weak}}\) & 27.316 & 0.265 & 0.623\\
        \(t_{\text{strong}}\) & \(t_{\text{strong}}\) & 5.528  & 0.353 & 0.620\\
        \midrule
        T        & 0        & 2.954  & 0.411 & 0.578\\
        \(t_{\text{weak}}\) & 0        & 2.923  & 0.404 & 0.576\\
        \(t_{\text{strong}}\) & 0        & 2.942  & 0.353 & 0.579\\
        \midrule
        T        & \(t_{\text{weak}}\) & 29.399 & 0.254 & 0.622\\
        T        & \(t_{\text{strong}}\) & 9.228  & 0.430 & 0.615\\
        \midrule
        \(t_{\text{weak}}\) & \(t_{\text{strong}}\) & 7.967  & 0.427 & 0.617\\
        \bottomrule
    \end{tabular}
    \caption{\noindent\textbf{Dual-Clock Ablation.}}
    \vspace{-10pt}
    \label{tbl:ablation}
\end{wraptable}

We ablate the dual-clock denoising scheme, presented in \ref{sec:dual_clock}, using the same evaluation protocol described in \ref{sec:obj-motion}.
In \shortname, the \emph{first tick} \(t_{\text{weak}}\) sets the initialization noise level for sampling, while the \emph{second tick} \(t_{\text{strong}}\) sets when we stop overriding the masked part with the noisy warped reference; after this point, all pixels denoise together.
For this section, we evaluate this procedure under different settings.
In these experiments, we use different timing ticks, denoting the first as \(t_1\) and the second as \(t_2\)

The resulting behaviors under different settings, together with their quantitative outcomes, are summarized below and in Table~\ref{tbl:ablation}:

\textbf{Single-clock baseline (\(t_1=t_2\)).}
  This implies applying SDEdit on the warped video (\(t_{\text{weak}}=t_{\text{strong}}\)).
  When \(t_1=t_2=t_{\text{weak}}\), too little conditioning is induced: the CoTracker distance is high, reflecting poor motion adherence.
  When \(t_1=t_2=t_{\text{strong}}\), non-masked regions become over-constrained to unintended motion, suppressing dynamics (e.g., the background freezes); see Fig.~\ref{Fig:dual_clock}, where the boat’s foam remains static although the boat moves.

\textbf{RePaint-style (\(t_2=0\)).}
  Here denoising occurs only outside the masked reference (equivalent to RePaint).
  As expected, for any \(t_1\) the CoTracker distance drops sharply, since the warped masked region is injected throughout denoising. 
  However, this comes at the cost of Imaging quality: the videos appear nearly perfect in motion adherence but unnatural overall, due to the lack of flexibility inside the mask region.

\textbf{Unconstrained background (\(t_1=T\)).}
  No constraint is applied to non-masked regions.
  For \(t_2=t_{\text{weak}}\), motion is not enforced and the model tends to generate overly static videos.
  For \(t_2=t_{\text{strong}}\), performance improves, but tracking error remains unsatisfactory; in practice, this setup often produces duplicate copies of the source object, which harms adherence.
  
\textbf{Dual clock (ours).}
\(t_1=t_{\text{weak}},\; t_2=t_{\text{strong}}\).
This setting achieves the best overall trade-off, combining strong motion-conditioning adherence (low CoTracker distance) with higher dynamic degree and robust visual quality.

\section{Implementation Details}\label{apdx:impl_details}

\subsection{Object Motion Control}\label{apdx:object_control}
This subsection complements Sec.~\ref{sec:obj-motion} with concise protocol and implementation details:

\begin{itemize}
\item \textbf{Single-Trajectory.}
To avoid ambiguity stemming from masks linked to multiple objects/trajectories in the original MC-Bench dataset, we restrict evaluation to single-trajectory cases (over 91\% of the dataset).
\item \textbf{Input handling.} Inputs are resized to each model’s native size and padded to match aspect ratio; after generation, padding is removed and outputs are resized back. Exceptions: MotionPro uses its original benchmark pipeline; DragAnything is run with its default input handling (we observed best results without external resizing).
\item \textbf{Trajectory scaling}: the 2D trajectory points are affinely transformed with the \emph{same} resize-and-pad mapping applied to the frames. After generation, we remove padding and invert the scaling when mapping tracks back for evaluation, ensuring geometric consistency.
\item \textbf{Clip length.} Standardized to 16 frames for SVD-based methods (as \cite{zhang2025motionpro}) and 49 frames for CogVideoX-based methods. 
Concretely, SVD emits 14 or 25 frames—thus \shortnamesvd generates 25 and keeps the first 16; DragAnything emits 20 and we keep the first 16; SG-I2V produces 14 and we evaluate the native 14-frame output, which may be slightly favorable to its metrics. 
If the output has more frames than the provided trajectory, we trim the trajectory to the target length.
\item \textbf{Pre/post-processing and prompts.} 
SG-I2V is conditioned on bounding boxes rather than masks, unlike the other methods. 
Therefore, following \citet{burgert2025go}, we supply the tight bounding box of the provided mask. 
For prompts, SVD-based methods are text-free, while CogVideoX-based methods use the MC-Bench prompts.
\item \textbf{Mask resizing}: 
Since both SVD and CogVideoX operate in a downsampled latent space, we project the binary masks to the latent resolution with nearest-neighbor interpolation. For SVD this is spatial-only; for CogVideoX also subsample in time to match temporal compression (nearest-neighbor in time).
\item \textbf{Hyperparameters.} All methods use $T{=}50$ denoising steps. For \shortnamesvd set $(t_{\text{weak}}, t_{\text{strong}})=(36,25)$ and fix MotionPro’s motion bucket to $17$ (as in their release). For \shortnamecog use $(46,41)$.
Other run-time settings follow each method’s defaults.
\item \textbf{VBench.} For CogVideoX-based 49-frame models, we use the long benchmark variant\footnote{\url{https://github.com/Vchitect/VBench/tree/master/vbench2_beta_long}}.
\item  \textbf{Dynamic Degree.}
VBench flags a clip as \emph{dynamic} when the mean of the top 5\% RAFT flow magnitudes in a frame exceeds a resolution-scaled threshold $\alpha\cdot\frac{\min(H,W)}{256}$ in at least 25\% of sampled frames. The default $\alpha\!=\!6.0$, tuned for VBench’s source videos, is too strict for our MC-Bench setting—predominantly static camera with small, localized motions—so nearly all clips are marked static. We therefore set $\alpha\!=\!3.5$ (keeping the 25\% rule unchanged), which yields a more meaningful separation of dynamic vs.\ static.
\item \textbf{CoTracker Distance (CTD).}
Following MC-Bench’s MD-Vid protocol, we leverage recent advances in point tracking~\citep{sand2008particle, tumanyan2024dino} and employ CoTracker~\citep{karaev2024cotracker} to assess the alignment between the input 2D trajectory and the motion in the generated video, measured by the mean per-frame Euclidean distance in pixel space (lower is better).
\item \textbf{Background–Object CoTracker Distance (BG--Obj CTD).}
This metric measures whether the background unintentionally moves together with the controlled object. 
We run CoTracker on the generated video for both (i) the tracked object trajectory and (ii) a uniform $16{\times}16$ grid of points sampled in the first frame. 
Let $o_t \in \mathbb{R}^2$ denote the object’s tracked position at frame $t$, and $p_{j,t} \in \mathbb{R}^2$ the tracked position of grid point $j$ at frame $t$. 
We convert all tracks to displacements from frame~1:
$\Delta o_t = o_t - o_1$, $\Delta p_{j,t} = p_{j,t} - p_{j,1}$. 
For each frame $t \ge 2$ and grid point $j$, we compute
$d_{j,t} = \lVert \Delta p_{j,t} - \Delta o_t \rVert_2$ (pixels). 
The BG--Obj CTD is then the average over frames and grid points:
\[
\mathrm{BG\text{--}Obj\ CTD} = \frac{1}{(T-1)J}\sum_{t=2}^{T}\sum_{j=1}^{J} d_{j,t}.
\]
Higher values indicate stronger object–background disentanglement (less co-motion).
\end{itemize}

\subsection{Camera Control on DL3DV}
\label{app:camera_control}
This subsection provides additional details for Sec. \ref{sec:cam-motion}.
For our camera control experiments, we use a subset of the DL3DV-10K dataset. The reference warped videos are created at a resolution of 960p using \href{https://pytorch3d.org/}{PyTorch3D}.

We utilize the official DL3DV camera transition data and align the coordinate systems with a sign flip of the z-axis and a flip of the camera pitch due to convention differences between PyTorch3D and NerfStudio, which was used to create the DL3DV dataset. The point cloud is generated in the original camera frame, with the camera extrinsics derived from parameter estimations and the estimated transition of the first frame. To resolve the inherent depth ambiguity, we perform a binary search to find the transition scale that maximizes the MSE alignment between the warped and original videos. This aligns the transitions to a metric scale consistent with the output of DepthPro.

To select a robust subset for evaluation, we first filter out videos from the 10K subset with an estimated scale of less than 0.3, as these were found to exhibit minimal real camera movement. We then select the 150 scenes with the lowest MSE loss between the warped and ground truth videos.
The strong guidance masks are generated by first marking pixels with no point cloud contribution as 0 (no guidance) and all other pixels as 1 (guidance). To ensure only regions with dense point cloud data are used for guidance, we apply a morphological "open" operation to the mask using an kernel size of 5. This operation serves to remove isolated noise and expand the non-guidance areas, resulting in a cleaner, more reliable mask.
For text guidance, we automatically generate a text prompt for each scene using GPT-4o~\citep{openai2024gpt4ocard}, following CogVideoX’s protocol\footnote{\url{https://github.com/zai-org/CogVideo/blob/main/inference/convert_demo.py}}.

\subsection{Appearance Control}
\label{app:app_control}
For the chameleon example demonstrating joint motion and appearance control, we use the prompt:
``A realistic video of a four-legged chameleon walking slowly and naturally from left to right along a thick, textured vine in a lush jungle. Its limbs move in a coordinated, controlled reptilian gait as it adjusts its body to the curve of the vine. The chameleon gradually changes color from green to purple.''

\section{Extra Qualitative Comparisons}
\label{apdx:qual_comp}
For the video versions of the comparisons in this paper, as well as additional results, please visit our \href{https://time-to-move.github.io/}{{project page}}.

\subsection{Qualitative Comparisons from MC-Bench}
Following the experiment described in Sec. \ref{sec:obj-motion}, we present additional results beyond those shown in Fig.~\ref{fig:object_drag_banchmark}, 
further illustrating our method's performance against leading approaches on the MC-Bench dataset using the SVD backbone.
\begin{figure}[h]
  \centering
  \includegraphics[width=1.\textwidth]{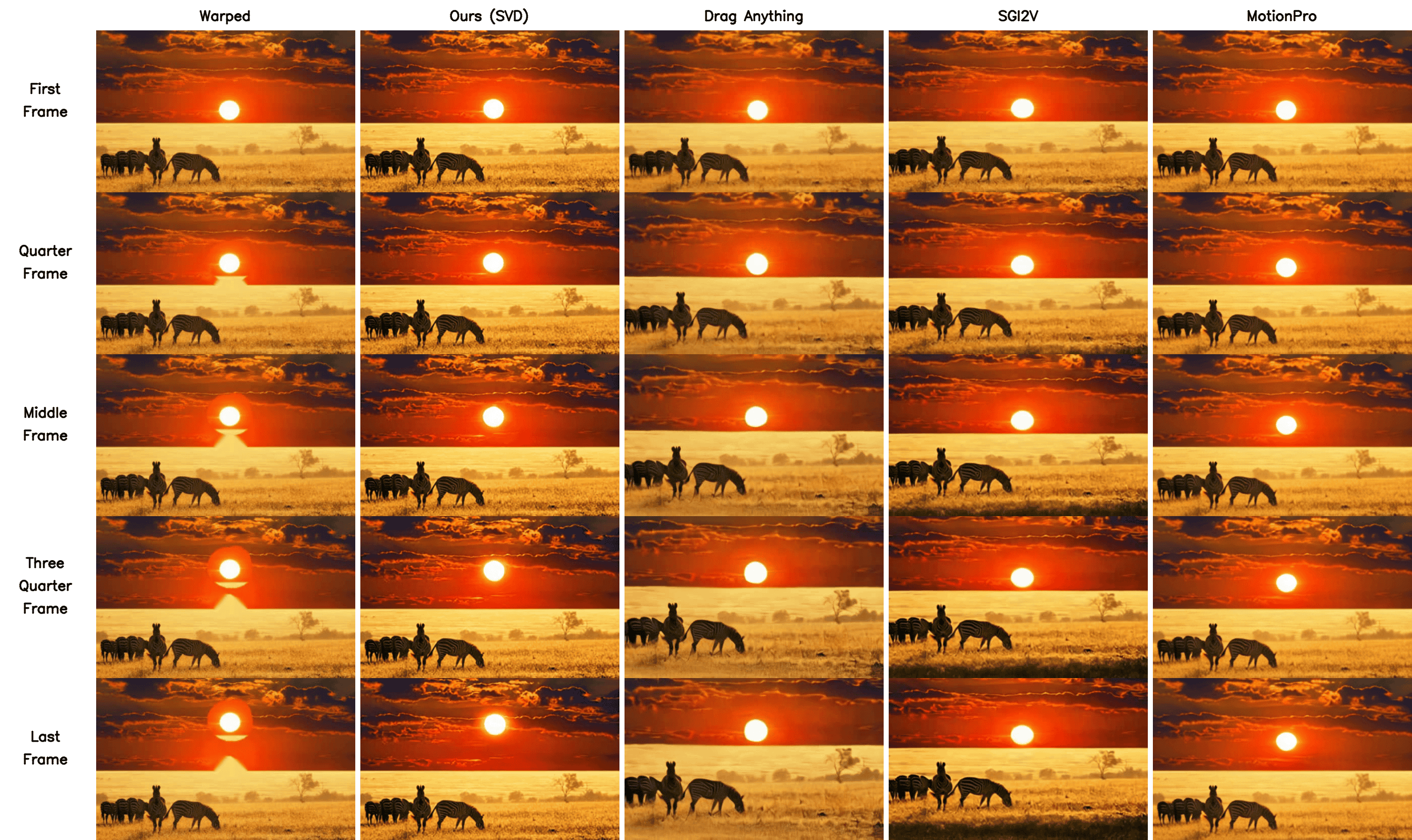}
  \label{fig:mc-bench-sunrise}
\end{figure}
\begin{figure}[h]
  \centering
  \includegraphics[width=1.\textwidth]{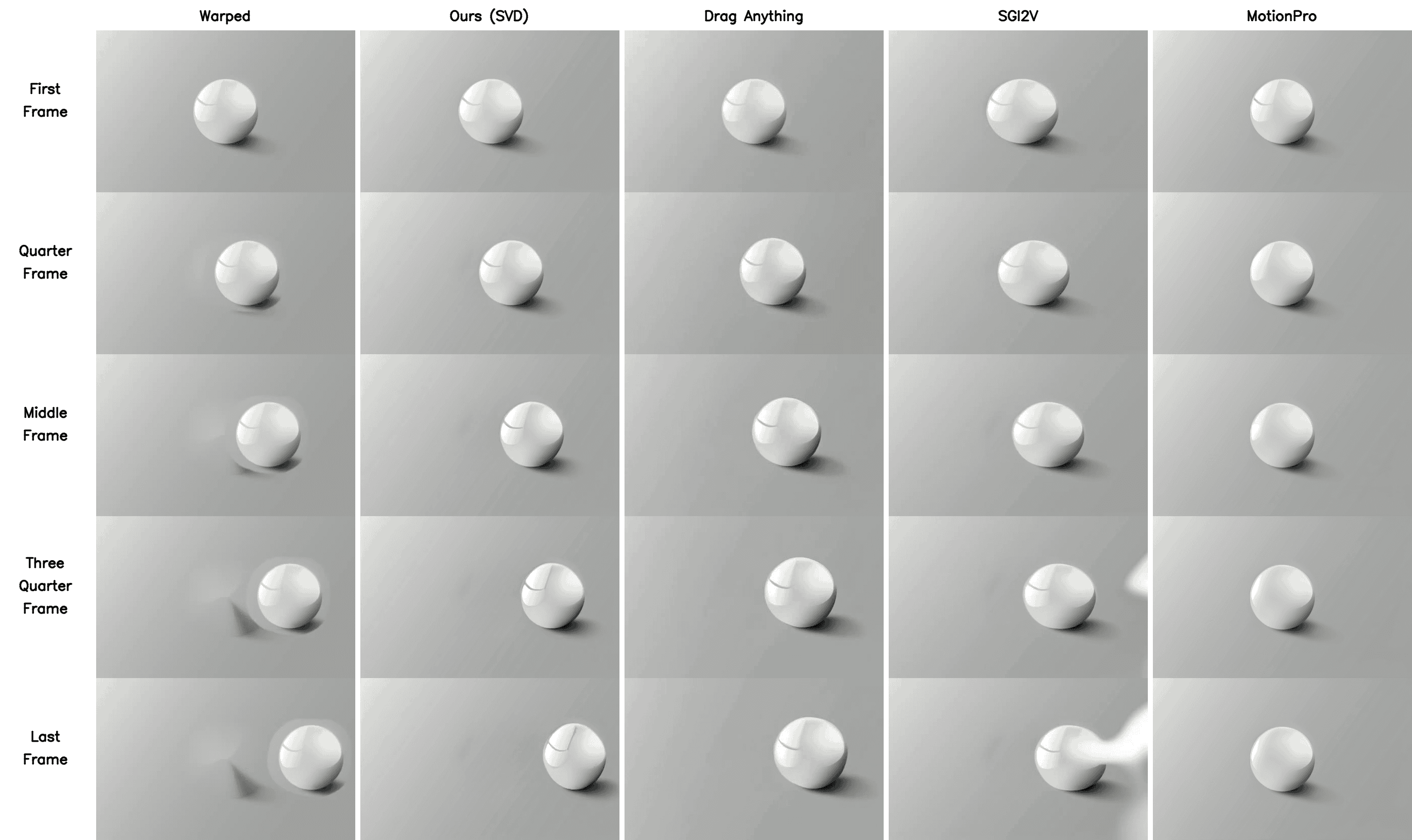}
  \label{fig:mc-bench-marble}
\end{figure}
\begin{figure}[h!] \centering
  \includegraphics[width=1.\textwidth]{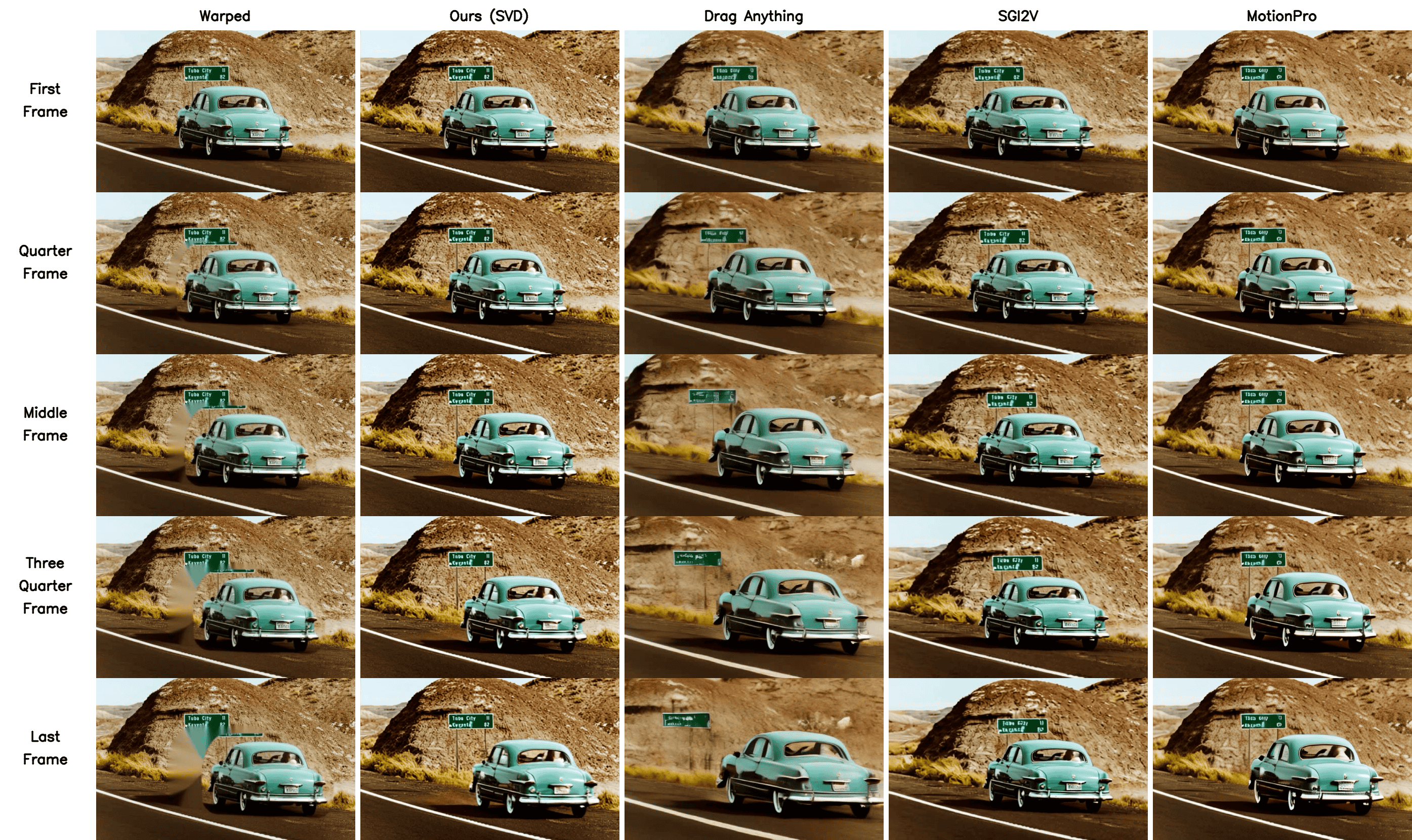} \label{fig:mc-bench-car} \end{figure}

\clearpage      
\subsection{Qualitative comparisons from DL3DV}
\label{apdx:qual_cam_ctrl}
Following the experiment described in Sec. \ref{sec:cam-motion}, we present qualitative results for camera-motion control on the DL3DV benchmark, comparing our method with GWTF given an input image, its monocular depth estimate, and a depth-warped video. 
These examples demonstrate superior performance in maintaining the intended camera motion and overall visual fidelity.
\begin{figure}[h!]
  \centering
  \includegraphics[width=1\textwidth]{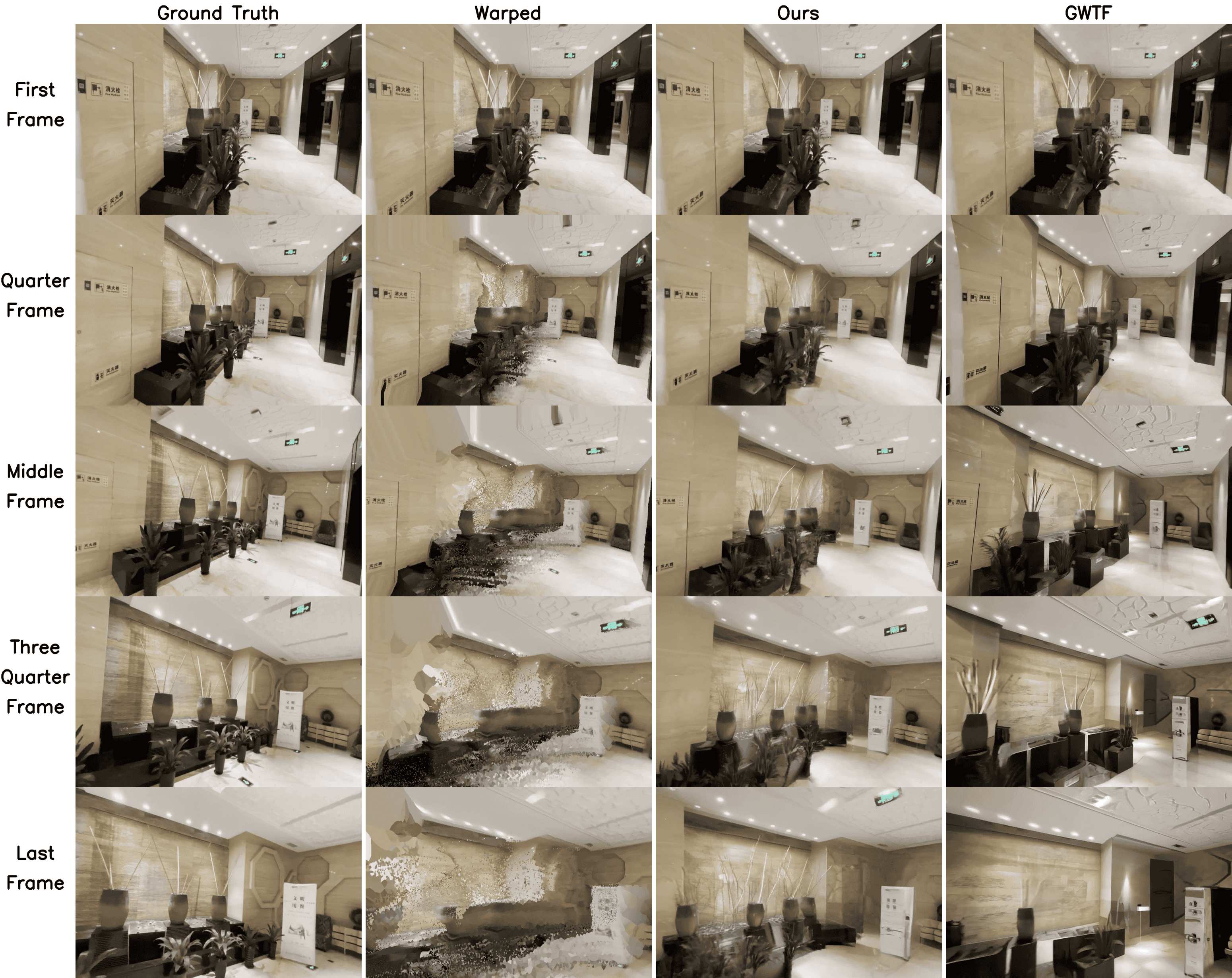}
  \label{fig:mc-bench-1}
\end{figure}
\begin{figure}[h!]
  \centering
  \includegraphics[width=1\textwidth]{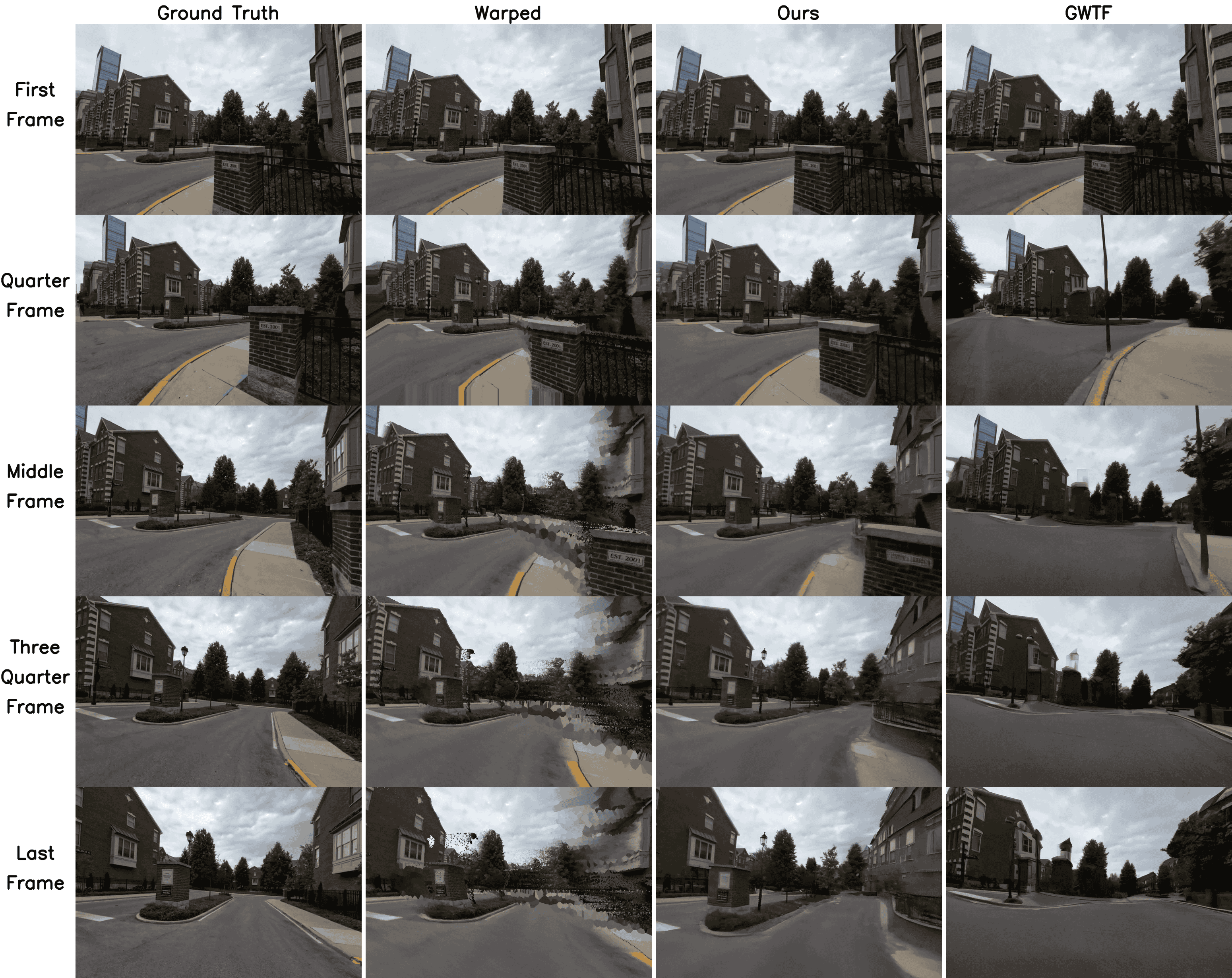}
  \label{fig:mc-bench-2}
\end{figure}
\begin{figure}[h!]
  \centering
  \includegraphics[width=1\textwidth]{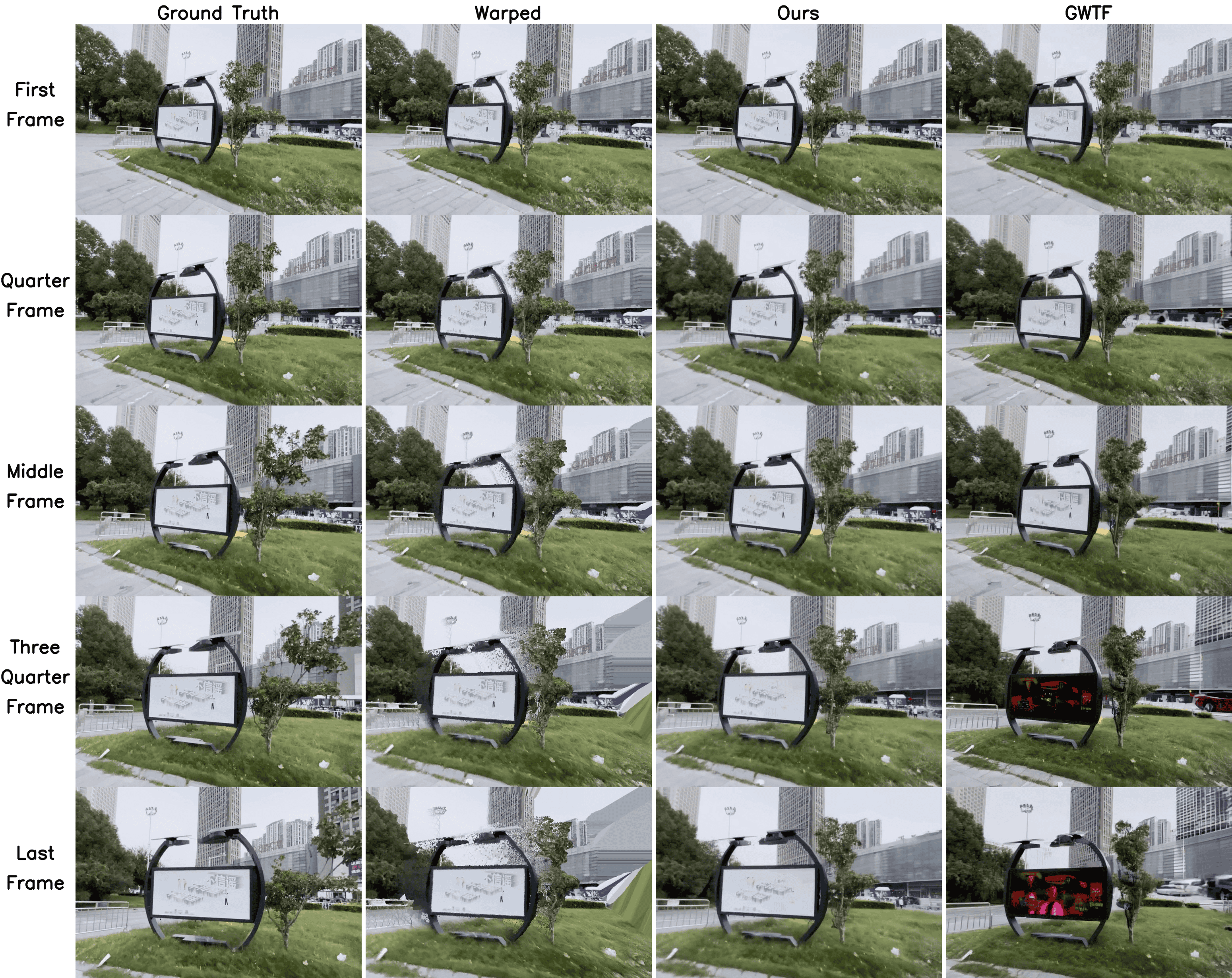}
  \label{fig:mc-bench-3}
\end{figure}

\clearpage      
\section{Challenging User-Created Examples}
\paragraph{Generation.}
To produce the examples shown in Fig.~\ref{fig:object_drag_human_crafted}, Fig. \ref{fig:camera_control_human_crafted}, and on the demo page, we collected 53 test cases, hand-crafted by users, spanning both object-motion and camera-motion control.
For each case, the initial reference frame was generated with Gemini~\citep{comanici2025gemini}, and object-control inputs were specified via a GUI adapted from the interface introduced in~\citep{burgert2025go}.
We will publicly release these examples at a later date.

\paragraph{Additional Results}
As explained in Sec. \ref{sec:we_wan}, we leverage the plug-and-play nature of our method to run on the recently released WAN2.2. 
Below we present additional “cut-and-drag” examples that complement Fig.~\ref{fig:object_drag_human_crafted}. 
These real-world cases illustrate scenarios in which the current state-of-the-art baseline, GWTF, often struggles to produce coherent results.

\begin{figure}[h!] \centering
  \includegraphics[width=1\textwidth]{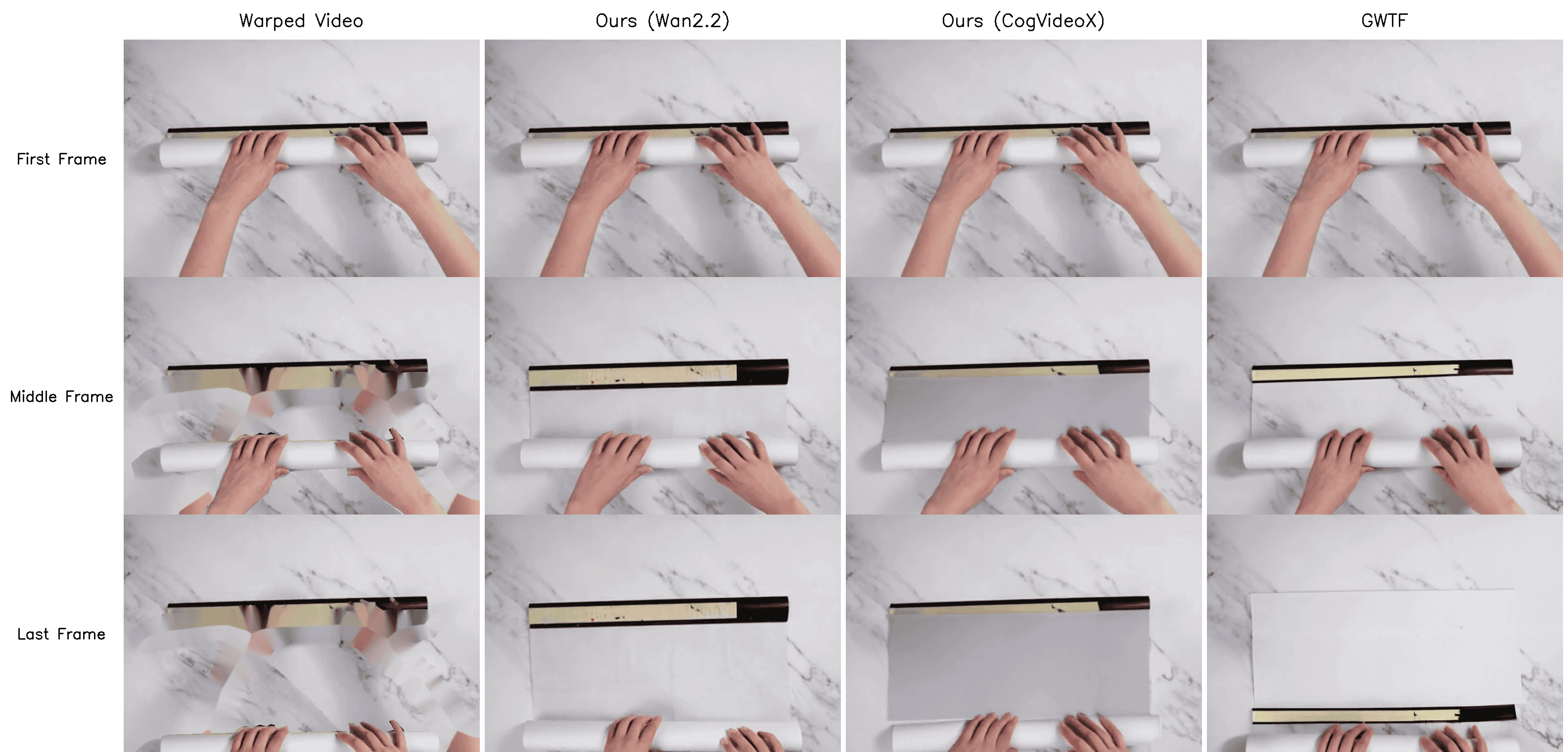} \label{fig:CutNDragExamples-PaperUnrolling} \end{figure}
\begin{figure}[h!] \centering
  \includegraphics[width=1\textwidth]{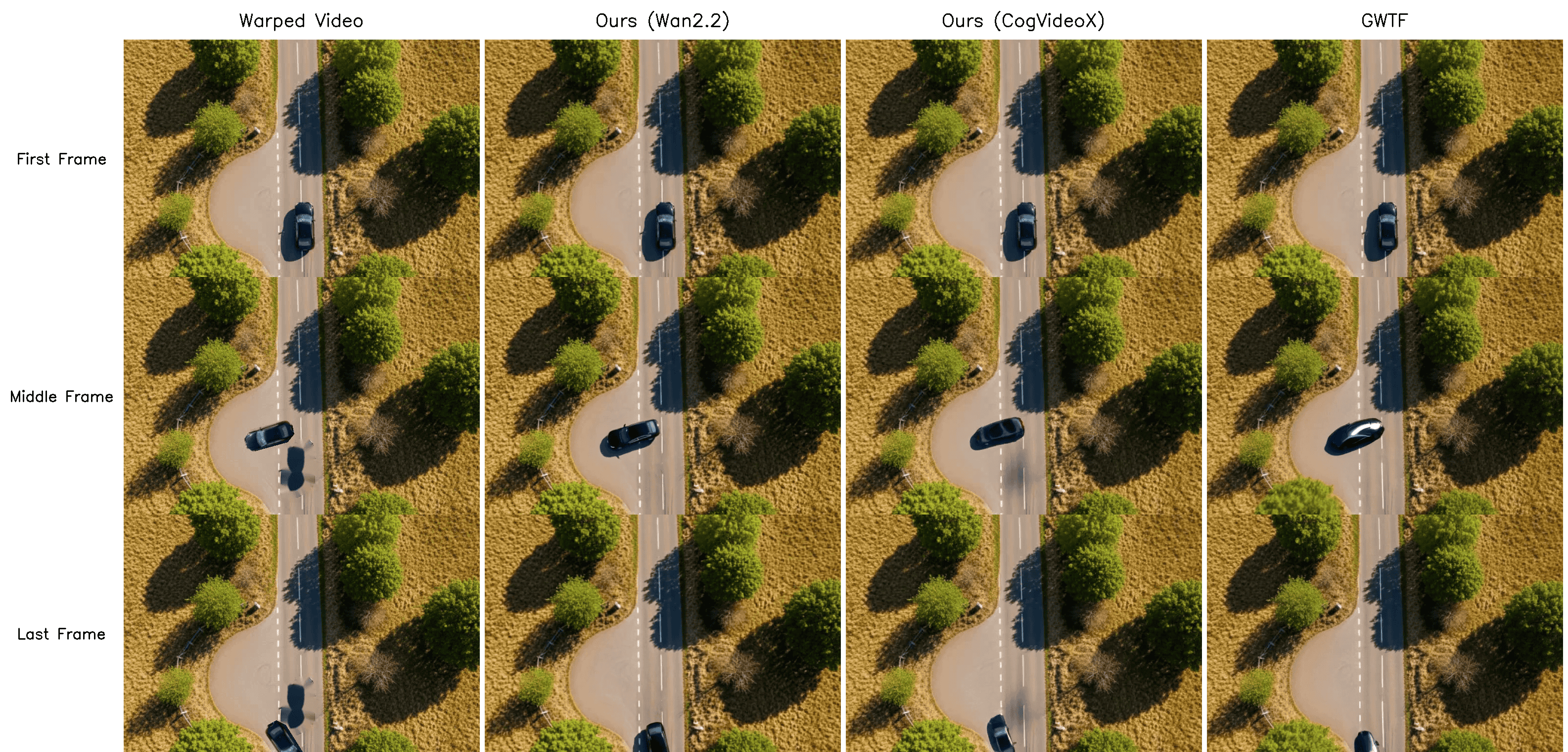} \label{fig:CutNDragExamples-CarUTurn} \end{figure}
\begin{figure}[h!] \centering
  \includegraphics[width=1\textwidth]{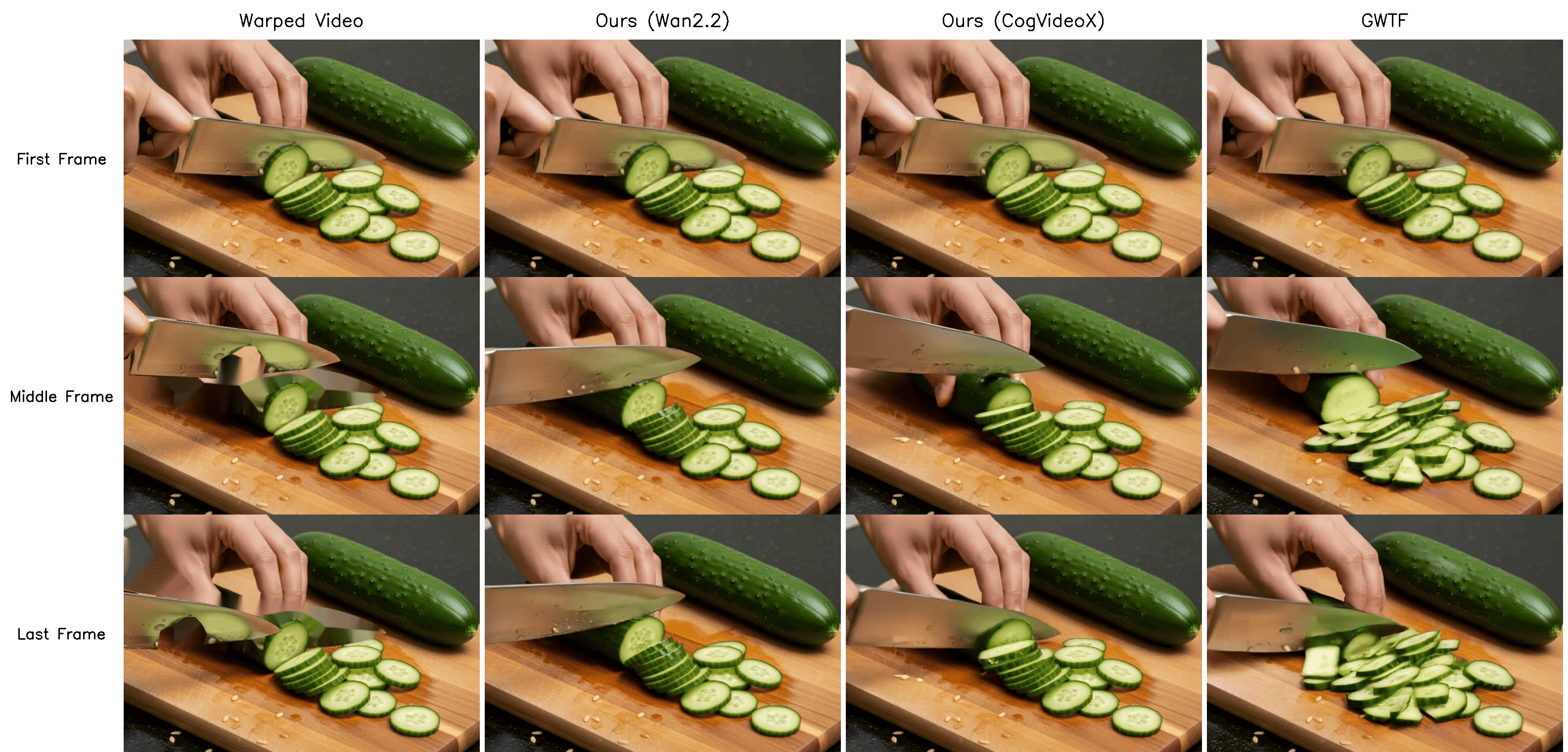} \label{fig:CutNDragExamples-CuttingCucumber} \end{figure}
\begin{figure}[h!] \centering
  \includegraphics[width=1\textwidth]{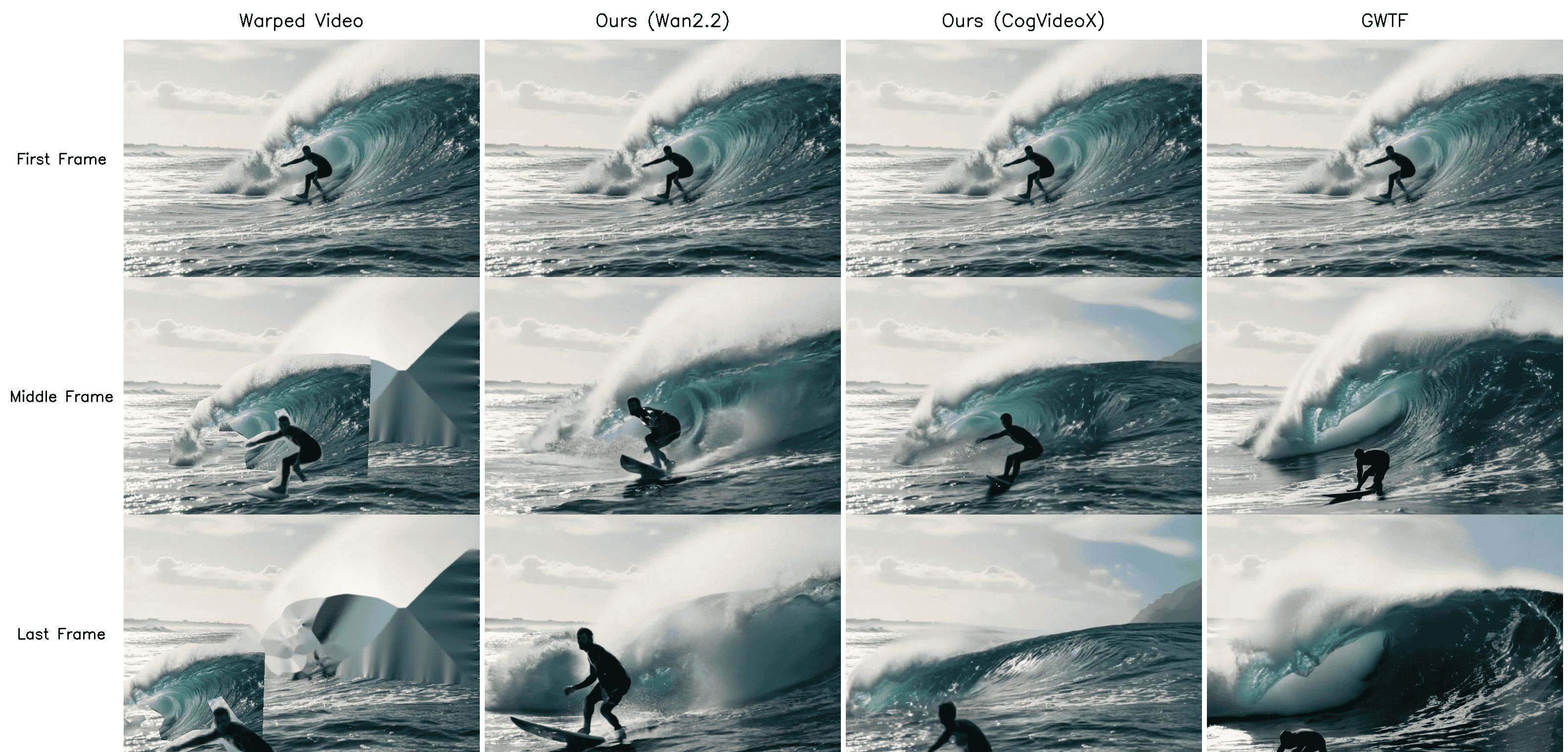} \label{fig:CutNDragExamples-Surfing} \end{figure}

\end{document}